\documentclass[lettersize,journal]{IEEEtran}
\usepackage{xcolor,soul,framed} 

\colorlet{shadecolor}{yellow}
\usepackage[pdftex]{graphicx}
\graphicspath{{../pdf/}{../jpeg/}}
\DeclareGraphicsExtensions{.pdf,.jpeg,.png}

\usepackage[cmex10]{amsmath}
\usepackage{bm}
\usepackage{array}
\usepackage{mdwmath}
\usepackage{mdwtab}
\usepackage{eqparbox}
\usepackage{url}
\usepackage{color} 
\usepackage{xcolor}
\usepackage{cite}
\usepackage{amssymb,amsfonts}
\usepackage{multirow}
\usepackage{float} 
\usepackage{algorithm}
\usepackage{algpseudocode}
\usepackage{booktabs}
\usepackage{lineno}
\usepackage{longtable}

\usepackage{enumitem}
\usepackage{makecell}

\usepackage[colorlinks,linkcolor=blue]{hyperref}

\usepackage{tcolorbox}
\definecolor{rq}{HTML}{1B365C}
\definecolor{rqBack}{HTML}{9ECBF7}

\usepackage[edges]{forest}
\usepackage{mydef}
\newcommand{\figref}[1]{Figure~\ref{#1}}

\newcommand{\secref}[1]{Section~\ref{#1}}

\usepackage{pifont}
\usepackage{threeparttable}

\newcommand{\gcmark}{\textcolor{green}{\checkmark}}
\newcommand{\rxmark}{\textcolor{red}{\ding{55}}}

\newcommand{\ie}{\textit{i}.\textit{e}.}
\newcommand{\eg}{\textit{e}.\textit{g}.}

\newtheorem{Def}{Definition}

\definecolor{hublink}{RGB}{247,106,184}

\usepackage{orcidlink}

\begin{document}


\title{Trajectory Data Management and Mining: \\A Survey from Deep Learning to the LLM Era}



\author{Wei Chen$^{\orcidlink{0009-0003-2260-9079}}$,Yuanshao Zhu$^{\orcidlink{0000-0002-5657-181X}}$,Yanchuan Chang$^{\orcidlink{0000-0002-1376-0311}}$,Kang Luo$^{\orcidlink{0000-0000-0000-0000}}$,Haomin Wen$^{\orcidlink{0000-0001-6130-126X}}$,Lei Li$^{\orcidlink{0000-0002-1386-767X}}$,Qingsong Wen$^{\orcidlink{0000-0003-4516-2524}}$,Yanwei Yu$^{\orcidlink{0000-0001-6941-2132}}$,\\Chao Chen$^{\orcidlink{0000-0003-2094-9734}}$,Kai Zheng$^{\orcidlink{0000-0002-0217-3998}}$,Yunjun Gao$^{\orcidlink{0000-0003-3816-8450}}$,Yu Zheng,~\IEEEmembership{Fellow,~IEEE}$^{\orcidlink{0009-0002-4515-0080}}$,Xiaofang Zhou,~\IEEEmembership{Fellow,~IEEE}$^{\orcidlink{0000-0001-6343-1455}}$,Yuxuan Liang$^{\dag}{\orcidlink{0000-0003-2817-7337}}$



\thanks{W.~Chen, Y.X~Liang, Y.S~Zhu, H.M~Wen, and L. Li are with Hong Kong University of Science and Technology (Guangzhou), Guangzhou, China. E-mail: onedeanxxx@gmail.com. Y.C~Chang is with The University of Melbourne, Australia. K.~Luo and Y.J~Gao are with Zhejiang University, Hangzhou, China. Y.W~Yu is with Ocean University of China, Qingdao, China. H.M~Wen is with Beijing Jiaotong University. Q.S~Wen is with Squirrel AI, USA. C.~Chen is with Chongqing University, Chongqing, China. K.~Zheng is with University of Electronic Science and Technology of China, Chengdu, China. W.~Chen and X.F~Zhou are with Hong Kong University of Science and Technology, Hongkong SAR. Y.~Zheng is with JD Intelligent Cities Research, JD Technology, Beijing, China. 
Y.X~Liang is the corresponding author. 
}

}



\markboth{Journal of \LaTeX\ Class Files,~Vol.~14, No.~8, August~2021}%
{Shell \MakeLowercase{\textit{et al.}}: A Sample Article Using IEEEtran.cls for IEEE Journals}


\maketitle

\begin{abstract}
Trajectory computing is a pivotal domain encompassing trajectory data management and mining, garnering widespread attention due to its crucial role in various practical applications such as location services, urban traffic, and public safety.
Traditional methods, focusing on simplistic spatio-temporal features, face challenges of complex calculations, limited scalability, and inadequate adaptability to real-world complexities. In this paper, we present a comprehensive review of the development and recent advances in trajectory computing, from deep learning to the more recent large language models. We first define trajectory data and provide a brief overview of widely-used deep learning models. Systematically, we explore deep learning applications in trajectory management (pre-processing, storage, analysis, and visualization) and mining (trajectory-related forecasting, trajectory-related recommendation, trajectory classification, travel time estimation, anomaly detection, and mobility generation). 
Furthermore, we discuss emerging research directions and recent advancements in large models (represented by foundation models and large language models) for trajectory computing, which promise to reshape the next generation of trajectory computing. Additionally, we summarize application scenarios, public datasets, and toolkits. Finally, we outline current challenges in trajectory computing research and propose future directions. Relevant papers and open-source resources have been collated and are continuously updated at: \textcolor{hublink}{\href{https://github.com/yoshall/Awesome-Trajectory-Computing}{https://github.com/yoshall/Awesome-Trajectory-Computing}}.
\end{abstract}



\section{Introduction}\label{sec:intro}
\IEEEPARstart{S}{ince} time immemorial, humanity has tirelessly attempted to study the science of mobility, driven by the fundamental laws that emerge from the micro and macro trajectory movements of objects~\cite{newton1687philosophiae,einstein1913entwurf,brockmann2006scaling}. The study of trajectories can be traced back as far as the 1960s. Researchers used various marking methods to track the movement trajectories of animals, discovering for the first time that movement behavior patterns possess geographical features and positivity among other patterns~\cite{sanderson1966study}. By the end of the 20th century, with the rapid development of Global Positioning System (GPS) and Geographic Information System technologies, it became possible to track spatial movement trajectories with long-term, high precision, and high efficiency. This includes volunteer positioning data, GPS-equipped travel trajectories, mobile terminal positioning, and communication records~\cite{zheng2010geolife}. These advancements have fueled the rise of trajectory research as a discipline, with wide-ranging applications in areas such as intelligent transportation, public safety, and business services~\cite{wen2023survey}.

\begin{figure}
    \centering
    \includegraphics[width=0.9 \linewidth]{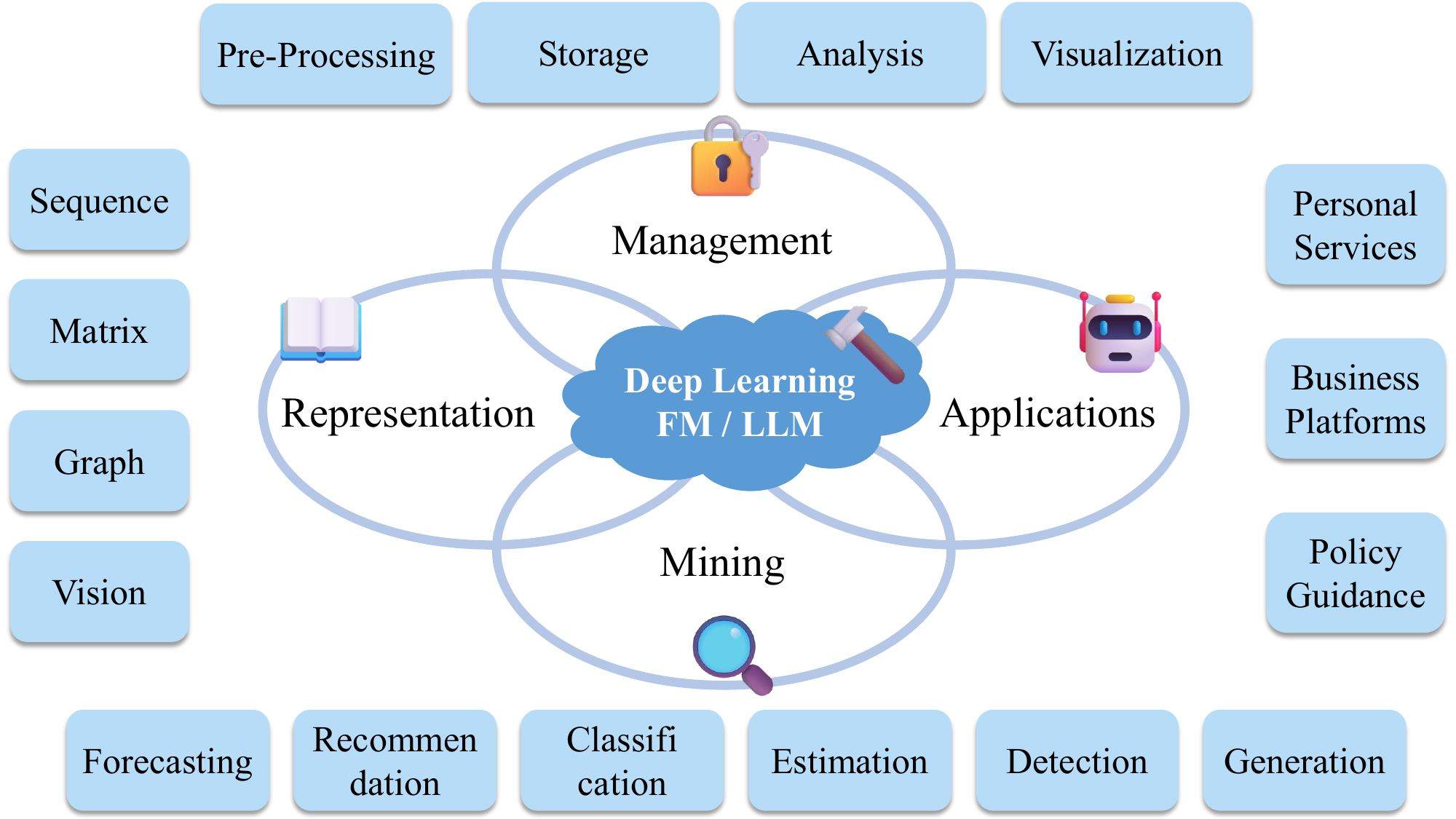}
    \caption{Trajectory computing overview.}
    \label{fig:traj overview}
    \vspace{-6mm}
\end{figure}

\begin{table*}[t!]
    \centering  
    \setlength{\tabcolsep}{1.9mm}{}
    \renewcommand\arraystretch{1.4}
    \caption{
        Comparison between this and other related surveys on data formats (i.e., sequence~(S), matrix~(M), graph~(G), and vision~(V)), relevant techniques (i.e., traditional methods~(TM), deep learning~(DL), and large language model~{LLM}), management tasks (i.e., pre-processing~(P), storage~(S),  analytics~(A), and visualization~(V)), and mining tasks (i.e., forecasting~(F), classification~(C), recommendation(R), estimation~(E), generation~(G), and detection~(D)). The number of downstream applications and publicly available datasets are also included. Besides, \gcmark~indicates content is covered, \rxmark~indicates that content is not covered, and $\color{hiddendraw}{\bm{\sim}}$ indicates that content is partially covered.
    }
    \vspace{-2mm}
    \scalebox{0.88}{
    \begin{tabular}{c|c|c|c|c|c|c|c|c|c|c|c|c|c|c|c|c|c|c|c|c}
    \toprule
    \multicolumn{1}{c}{\multirow{2}{*}{\textbf{Survey}}} & \multicolumn{1}{c|}{\multirow{2}{*}{\textbf{Year}}} & \multicolumn{4}{c|}{\textbf{Formats}} & \multicolumn{3}{c|}{\textbf{Techniques}} & \multicolumn{4}{c|}{\textbf{Management}} & \multicolumn{6}{c|}{\textbf{Mining}} & \multicolumn{1}{c}{\multirow{2}{*}{\textbf{\#Applications}}} & \multicolumn{1}{c}{\multirow{2}{*}{\textbf{\#Public Datasets}}} \\ 
    \cline{3-6} \cline{7-9} \cline{10-13} \cline{14-19}
    \multicolumn{2}{c|}{} & \textbf{S} & \textbf{M} & \textbf{G} & \textbf{V} & \textbf{TM} & \textbf{DL} & \textbf{LLM} & \textbf{P} & \textbf{S} & \textbf{A} & \textbf{V} & \textbf{F}  & \textbf{R} & \textbf{C} & \textbf{E} & \textbf{G} & \textbf{D} & \multicolumn{2}{c}{} \\
    \hline
    
    Zheng \textit{et al.}~\cite{zheng2015trajectory} & 2015 & \gcmark & \gcmark & \gcmark & \rxmark & \gcmark & \rxmark & \rxmark & \gcmark & \gcmark & \gcmark & \rxmark & \rxmark & \gcmark & \gcmark & \gcmark & \rxmark & \gcmark & 6 & 10 \\

    Feng \textit{et al.}~\cite{feng2016survey} & 2016 & \gcmark & \rxmark & \gcmark & \rxmark & \gcmark & \rxmark & \rxmark & \gcmark & \gcmark & \gcmark & \rxmark & $\color{hiddendraw}{\bm{\sim}}$ & $\color{hiddendraw}{\bm{\sim}}$ & $\color{hiddendraw}{\bm{\sim}}$ & \rxmark & \rxmark & \rxmark & 6 & \rxmark \\

    Mazimpaka \textit{et al.}~\cite{mazimpaka2016trajectory} & 2016 & \gcmark & \gcmark & \gcmark & \rxmark & \gcmark & \rxmark & \rxmark  & \rxmark & \rxmark & $\color{hiddendraw}{\bm{\sim}}$ & \rxmark & \gcmark & \gcmark & \gcmark & \rxmark & \rxmark & \gcmark & 13 & \rxmark \\
    
    Bian \textit{et al.}~\cite{bian2018survey} & 2018 & \gcmark & \gcmark & \rxmark & \gcmark & \gcmark & \gcmark  & \rxmark & $\color{hiddendraw}{\bm{\sim}}$ & \rxmark & \gcmark & \rxmark & \rxmark & \rxmark & \rxmark & \rxmark & \rxmark & \rxmark & 7 & \rxmark \\

    Bian \textit{et al.}~\cite{bian2019trajectory} & 2019 & \gcmark & \gcmark & \rxmark & \gcmark & \gcmark & \gcmark  & \rxmark & $\color{hiddendraw}{\bm{\sim}}$ & \rxmark & \gcmark & \rxmark & \rxmark & \rxmark & \gcmark & \rxmark & \rxmark & \rxmark & 7 & 6 \\

    Koolwal \textit{et al.}~\cite{koolwal2020comprehensive} & 2020 & \gcmark & \gcmark & \gcmark & \rxmark & \gcmark & \gcmark  & \rxmark & \gcmark & \rxmark & \gcmark & \rxmark & \gcmark & $\color{hiddendraw}{\bm{\sim}}$ & $\color{hiddendraw}{\bm{\sim}}$ & \rxmark & \rxmark & \rxmark & 9 & 18 \\

    Wang \textit{et al.}~\cite{wang2021survey} & 2021 & \gcmark & \gcmark & \gcmark & \rxmark & \gcmark & \gcmark  & \rxmark & \gcmark & \gcmark & \gcmark & \rxmark & \gcmark & $\color{hiddendraw}{\bm{\sim}}$ & \gcmark & \gcmark & $\color{hiddendraw}{\bm{\sim}}$ & $\color{hiddendraw}{\bm{\sim}}$ & 13 & 20 \\


    Luca \textit{et al.}~\cite{luca2021survey} & 2021 & \gcmark & \gcmark & \gcmark & \rxmark & \gcmark & \gcmark  & \rxmark & \rxmark & \rxmark & \rxmark & \rxmark & \gcmark & \rxmark & \rxmark & \rxmark & \gcmark & \rxmark & 7 & 18 \\
    
    Aghababa \textit{et al.}~\cite{pourmahmood2022classifying} & 2022 & \gcmark & \rxmark & \rxmark & \gcmark & \gcmark & \gcmark  & \rxmark & \gcmark & \rxmark & \rxmark & \rxmark & \rxmark & \rxmark & \gcmark & \rxmark & \rxmark & \rxmark & \rxmark & 6 \\

    Shaygan \textit{et al.}~\cite{shaygan2022traffic} & 2022 & \gcmark & \gcmark & \gcmark & \rxmark & \gcmark & \gcmark  & \rxmark & \gcmark & \gcmark & \rxmark & \rxmark & \gcmark & \rxmark & \rxmark & \gcmark & \rxmark & \gcmark & 6 & 18 \\

    Duarte \textit{et al.}~\cite{duarte2023benchmark} & 2023 & \gcmark & \rxmark & \rxmark & \rxmark & \gcmark & \gcmark  & \rxmark & \rxmark & \rxmark & \rxmark & \rxmark & \rxmark & \rxmark & \rxmark & \rxmark & \rxmark & \gcmark & \rxmark & 4 \\
    
    Hu \textit{et al.}~\cite{hu2023spatio} & 2023 & \gcmark & \gcmark & \gcmark & \rxmark & \gcmark & \gcmark  & \rxmark & $\color{hiddendraw}{\bm{\sim}}$ & $\color{hiddendraw}{\bm{\sim}}$ & \gcmark & \rxmark & \rxmark & \rxmark & \rxmark & \rxmark & \rxmark & \rxmark & \rxmark & 4 \\
    
    
    Graser \textit{et al.}~\cite{graser2024mobilitydl} & 2024 & \gcmark & \gcmark & \gcmark & \gcmark & \rxmark & \gcmark  & \rxmark & \rxmark & \rxmark & \rxmark & \rxmark & \gcmark & \rxmark & \gcmark & $\color{hiddendraw}{\bm{\sim}}$ & \gcmark & \gcmark & 8 & 17 \\
    
    \hline

    \textbf{This Survey} & 2025 & \gcmark & \gcmark & \gcmark & \gcmark & \gcmark & \gcmark  & \gcmark & \gcmark & \gcmark & \gcmark & \gcmark & \gcmark & \gcmark & \gcmark & \gcmark & \gcmark & \gcmark & 15 & 38 \\ \bottomrule

    \end{tabular}
}
    \vspace{-6mm}
\label{table:compare_survey}
\end{table*}

However, the effective management and mining of vast records of highly refined trajectories and quantitative spatio-temporal distribution data presents an urgent challenge. Over the past two decades, extensive research has led to a comprehensive framework and theory for trajectory computing, encompassing the entire analysis process: pre-processing (\eg, map matching, stay point detection~\cite{zheng2011computing}), indexing and retrieval (\eg, similarity linking, regional / semantic querying~\cite{shang2017trajectory,wang2021survey}), pattern mining, and uncertainty modeling~\cite{zheng2008understanding,zheng2011computing}. Despite numerous efficient, stage-specific algorithms developed for these loosely coupled processes, three key challenges persist: 1) \textit{Lack of uniformity.} Problem modeling remains difficult due to the need to combine various tools (\eg, rule-based and probabilistic) depending on the scenario. 2) \textit{Complexity.} The inherent spatio-temporal heterogeneity and auto-correlation in raw trajectory data complicate the capture of intrinsic features via feature engineering or simple expert rules. 3) \textit{Adaptability.} Traditional technologies often suffer from the curse of dimensionality when processing massive data and struggle to adapt to new application scenarios.


In recent years, we have witnessed the rapid rise of deep learning in various fields~\cite{lecun2015deep}, attributed to its remarkable end-to-end modeling and representation capabilities. Beyond conventional data types (\eg, text, images, audio), it has also been extended to general spatio-temporal data with irregular structures \cite{wang2019deep}. Trajectory data, characterized by its integrated spatial, temporal, and semantic dimensions, represents a typical case of such data. Accordingly, researchers have leveraged deep learning to reconstruct core components of the trajectory computing framework, including efficient trajectory data management \cite{wang2021survey}, effective trajectory data mining \cite{luca2021survey}, and various novel downstream applications \cite{duarte2023benchmark}. By virtue of diverse neural network architectures and learning paradigms, traditional trajectory-related problems are converted into learning tasks in a seamless manner. Moreover, the integration of prior expert knowledge from spatial statistics, geometry, and geography enables these models to capture complex spatio-temporal trajectory patterns, thereby facilitating the development of innovative applications. In Fig.~\ref{fig:traj overview}, we provide an overview of trajectory computing.

\textbf{Related Surveys.} While deep learning is increasingly utilized for trajectory computing, existing surveys often have a limited scope, focusing individually on aspects like trajectory management (\eg, clustering~\cite{bian2018survey,chen2019real}, similarity~\cite{hu2023spatio}, privacy~\cite{jin2022survey}) or mining (\eg, location prediction~\cite{koolwal2020comprehensive,jiang2018survey}, recommendation~\cite{safavi2022toward}, arrival time estimation~\cite{reich2019survey,wen2023survey}), and only partially address deep learning techniques. Surveys on spatio-temporal data mining~\cite{wang2020deep,jin2023large,gao2022generative} and intelligent traffic~\cite{veres2019deep,yuan2021survey} are also prevalent but offer limited coverage of trajectory content. Notably, recent surveys on deep learning for trajectory data mining~\cite{graser2024mobilitydl,luca2021survey} neglect trajectory data management. Furthermore, the integration of nascent large models (\eg, LLMs~\cite{achiam2023gpt}) with trajectory tasks is emerging~\cite{jin2023spatio}, but lacks a dedicated review. These limitations highlight the urgent need for a comprehensive review, a distinction summarized in Tab.~\ref{table:compare_survey}.

\textbf{Our Contributions.} To address the literature gap, this study offers a systematic and up-to-date review of deep learning and large models for trajectory computing, summarized as follows:
\begin{itemize}[leftmargin=*]
    \item \textbf{First Systematic Survey.} This work provides the first comprehensive review of deep learning advances in trajectory computing, notably highlighting cutting-edge progress with large models, including Foundation Models (FM) and Large Language Models (LLM), offering a thorough overview.
    \item \textbf{Unified and Structured Taxonomy.} We propose a unified taxonomy that structures trajectory computing into three parts: elaborating on diverse trajectory data forms, identifying common tasks in trajectory management and mining, and presenting practical applications across multiple domains. This classification facilitates systematic understanding.
    \item \textbf{Comprehensive Resource Collection.} We initiate the Trajectory Computing Project, an open-sourced, continuously updated repository curating the most comprehensive collection of trajectory datasets, resources, and academic papers on deep learning and large models for trajectory management and mining, serving researchers, engineers, and urban planners.
    \item \textbf{Future Directions and Opportunities.} We analyze the latest advances in large model-enhanced trajectory computing (\eg, FM, LLM) amid the ongoing paradigm shift and outline several promising future research directions, providing guidance for the field's development.
\end{itemize}

\vspace{-2mm}
\section{Preliminary}~\label{sec:pre}
\vspace{-8mm}
\subsection{Definition and Notation}~\label{sec:defi_nota}
\vspace{-4mm}
\begin{Def}[Spatio-Temporal Point]
A spatio-temporal point $p$ is a unique entity in the form of $(o,t,l,f)$, where represents the access of moving object $o$ to location $l$ at timestamp $t$ under the geographical coordinate system, and comes with an optional record attribute feature $f$.
\end{Def}
\begin{Def}[Trajectory] 
A generalized trajectory $T$ consists of a series of spatial-temporal point sequences $(p_1,p_2,...,p_n)$ arranged in chronological order, which represents the movement information generated by moving objects in geographical space.
\end{Def}

Based on the fundamental attributes of spatio-temporal points, trajectory can be extended into various forms. Firstly, with respect to object attributes, we can categorize them into \textbf{\textit{Individual Trajectory}}, representing quasi-continuous tracking data of individual movements, and \textbf{\textit{Group Trajectory}}, which denote the movements of a group of individuals during the observation period, typically aggregated into edges/nodes, grids, or a set of Points of Interest (POI) in the mobility graph. Secondly, regarding time attributes, we can derive a spectrum of trajectories ranging from \textbf{\textit{Sparse Trajectory}} (\eg, users' check-in data during travel) to \textbf{\textit{Dense Trajectory}} (\eg, movement paths of vehicles equipped with GPS tracking systems) based on the dimension of sampling frequency. Thirdly, regarding location attributes, we can generate trajectories, also known as \textbf{\textit{Raw Trajectory}}, by mapping coordinates to spatial embeddings to discretize the geographic space system. The newly generated sequence of discretized tokens is referred to as \textbf{\textit{Cell Trajectory}}. Further, trajectories composed of tokens with attribute features are termed as \textbf{\textit{Semantic Trajectory}}. The relationship of all the above attributes of trajectories is illustrated in Fig.~\ref{fig:traj example}.

\vspace{-3mm}
\subsection{Unique Properties of Trajectory Data} 


Trajectory data exhibits unique characteristics that are pivotal for understanding spatial-temporal movements and predicting urban mobility patterns. The following properties underscore the complexity and richness of trajectory data:

\begin{figure}[t!]
    \centering
    \includegraphics[width=0.9\linewidth]{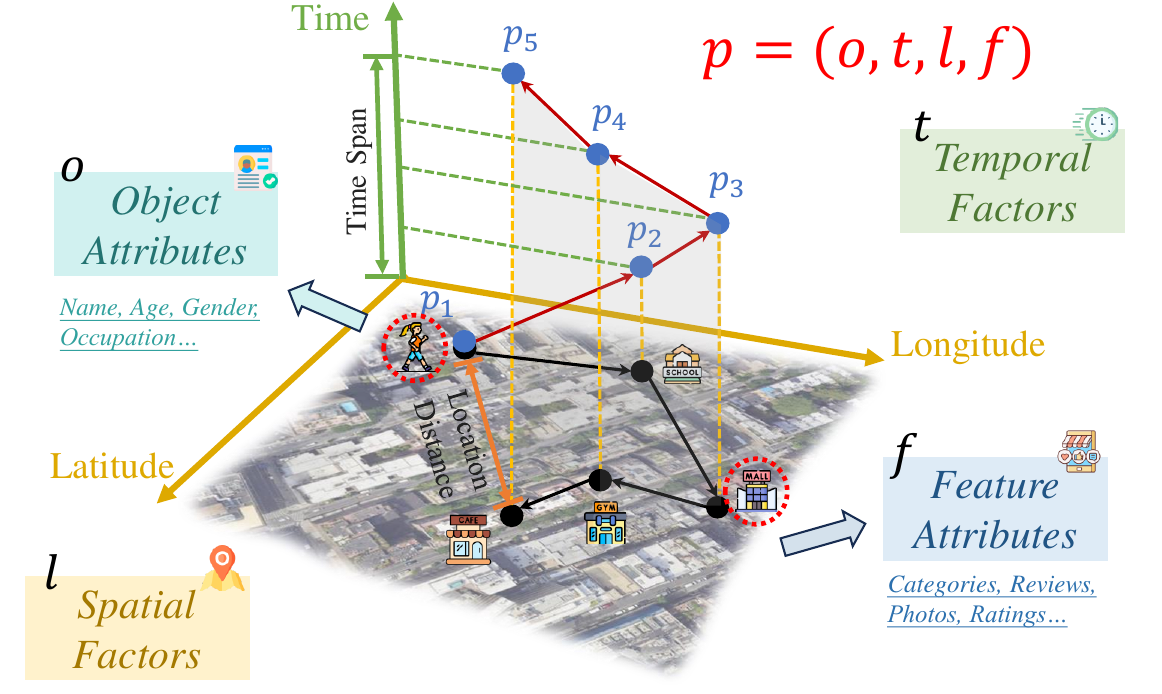}
    \caption{Illustration of a trajectory.}
    \label{fig:traj example}
    \vspace{-6mm}
\end{figure}


\begin{itemize}[leftmargin=*] 
    \item \textbf{Spatio-temporal dependencies}. Trajectory data inherently exhibits spatio-temporal dependencies as a sequence of spatial locations over time. These dependencies reveal high-level patterns of transfer modes and travel intentions, which are crucial for movement behavior analysis and forecast. 
    \item \textbf{Personalization}. As trajectory data is generated by specific individuals or entities, it contains personalized traits reflecting subjects' preferences and mobility habits. Accurate modeling of these personalized features is imperative for enhancing the precision of micro-level traffic behavior prediction tasks. 
    \item \textbf{Irregularity}. Trajectory data often suffers from irregularity due to sampling limitations or data compression. This property results in insufficient supervisory information—e.g., missing detailed path information between points—which poses a significant challenge and can degrade performance in movement prediction tasks.
\end{itemize}

Each of these properties contributes to the complexity of handling trajectory data, demanding sophisticated modeling techniques to accurately interpret and predict mobility patterns in urban computing contexts.

\tikzstyle{leaf}=[draw=hiddendraw,
    rounded corners, minimum height=1em,
    fill=myblue!40,text opacity=1, 
    fill opacity=.5,  text=black,align=left,font=\scriptsize,
    inner xsep=3pt,
    inner ysep=1pt,
    ]
\tikzstyle{middle}=[draw=hiddendraw,
    rounded corners, minimum height=1em,
    fill=output-white!40,text opacity=1, 
    fill opacity=.5,  text=black, align=center,font=\scriptsize,
    inner xsep=7pt,
    inner ysep=1pt,
    ]

\begin{figure*}[t!]
\centering
\scalebox{0.78}
{
\begin{forest}
  for tree={
      forked edges,
      grow=east,
      reversed=true,
      anchor=base west,
      parent anchor=east,
      child anchor=west,
      base=middle,
      font=\scriptsize,
      rectangle,
      line width=0.7pt,
      draw=output-black,
      rounded corners,align=left,
      minimum width=2em, s sep=6pt, l sep=8pt,
      where level=1{text width=0.2\linewidth}{},
      where level=2{text width=0.2\linewidth,font=\scriptsize}{},
      where level=3{font=\scriptsize}{},
      where level=4{font=\scriptsize}{},
      where level=5{font=\scriptsize}{}
  },
  [Taxonomy, middle, rotate=90, anchor=north, edge=output-black, 
      [{Trajectory Data Management\\ (\secref{sec:management})},middle,anchor=west,edge=output-black, text width=0.18\linewidth, 
        [{Pre-Processing\\ (\secref{subsec:pre-processing})}, middle, text width=0.15\linewidth, edge=output-black,
            [Simplification, middle, text width=0.1\linewidth, edge=output-black,
                [{RLTS~\cite{trajsimp_zhengwang_rl}, S3~\cite{trajsimp_lightweight}, MARL4TS~\cite{trajsimp_errorbound}, RL4QDTS~\cite{wang2023collectively}}, leaf, text width=0.43\linewidth, edge=output-black]
            ],
            [Recovery, middle, text width=0.1\linewidth, edge=output-black,
                [{DHTR~\cite{wang2019deep}, AttnMove~\cite{xia2021attnmove}, PeriodicMove~\cite{sun2021periodicmove}, TrajBERT~\cite{si2023trajbert}, \\TEIR~\cite{chen2023teri}, MTrajRec~\cite{ren2021mtrajrec}, RNTrajRec~\cite{chen2023rntrajrec}, VisionTraj~\cite{li2023visiontraj}, \\STR~\cite{long2024learning}, Traj2Traj~\cite{liao2023traj2traj}, PATR~\cite{zhang2022patr}, DeepMG \cite{ruan2020learning}, DelvMap~\cite{wang2024delvmap}}, leaf, text width=0.43\linewidth, edge=output-black]
            ],
            [Map-Matching, middle, text width=0.1\linewidth, edge=output-black,
                [{DeepMM~\cite{mm-deepmm}, \cite{mm-transformer}, L2MM~\cite{jiang2023l2mm}, GraphMM~\cite{mm-graph}, \\DMM~\cite{mm-dmm}, TBMM~\cite{zhu2023map}, FL-AMM~\cite{mm-fl}}, leaf, text width=0.43\linewidth, edge=output-black]
            ]
        ],
        [{Storage\\ (\secref{subsec:storage})}, middle, text width=0.15\linewidth, edge=output-black,
            [Storage Database, middle, text width=0.1\linewidth, edge=output-black,
                [{TrajMesa~\cite{li2021trajmesa}, Milvus~\cite{wang2021milvus}, ITS-DB~\cite{cai2017vector}, Ghost~\cite{fang2023ghost}}, leaf, text width=0.43\linewidth, edge=output-black]
            ],
            [Index \& Query, middle, text width=0.1\linewidth, edge=output-black,
                [{TraSS~\cite{he2022trass}, \cite{qi2020effectively}, ~\cite{pandey2020case}, X-FIST~\cite{ramadhan2022x}}, leaf, text width=0.43\linewidth, edge=output-black]
            ]
        ],
        [{Analytics\\ (\secref{subsec:analytics})}, middle, text width=0.15\linewidth, edge=output-black,
            [Similarity Measurement, middle, text width=0.15\linewidth, edge=output-black, 
                [{RSTS~\cite{trajsimi_rsts}, Tedj.~\cite{trajsimi_tedjopurnomo}, At2vec~\cite{trajsimi_at2vec}, CL-Tsim~\cite{trajsimi_cltsim}, TrajCL~\cite{trajsimi_trajcl}, \\TrajRCL~\cite{li2023self}, CSTRM~\cite{trajsimi_cstrm}, TrjSR~\cite{trajsimi_trjsr}, Play2vec~\cite{trajsimi_play2vec}, \\NEUTRAJ~\cite{trajsimi_neutraj}, Traj2SimVec~\cite{trajsimi_traj2simvec}, \cite{trajsimi_chen_iotj}, TMN~\cite{trajsimi_tmn}, T3S~\cite{trajsimi_t3s}, \\TrajGAT~\cite{trajsimi_trajgat}, Trembr~\cite{trajsimi_trembr}, LightPath~\cite{yang2023lightpath}, GTS~\cite{trajsimi_gts}, \\GRLSTM~\cite{trajsimi_grlstm}, GTS+~\cite{zhou2023spatial}, ST2Vec~\cite{trajsimi_st2vec}, SARN~\cite{trajsimi_sarn}}, leaf, text width=0.38\linewidth, edge=output-black]
            ],
            [Cluster Analysis, middle, text width=0.15\linewidth, edge=output-black,
                [{Trip2Vec~\cite{trajclus_trip2vec}, \cite{trajclus_olive2020deep}, DETECT~\cite{trajclus_DETECT}, E2DTC~\cite{trajclus_E2DTC}}, leaf, text width=0.38\linewidth, edge=output-black]
            ]
        ],
        [{Visualization\\ (\secref{subsec:visualization})}, middle, text width=0.15\linewidth, edge=output-black,
            [{DeepHL~\cite{maekawa2020deep}, Surveillance\cite{lee2019visual}, DSAE\cite{liu2017visualization},\cite{zhou2018visual}, \cite{zhang2021deep}}, leaf, text width=0.575\linewidth, edge=output-black]
        ]
    ],
    [{Trajectory Data Mining\\ (\secref{sec:mining})}, middle,anchor=west,edge=output-black, text width=0.18\linewidth,
        [{Trajectory-related\\Forecasting\\(\secref{subsec:forecasting})}, middle, text width=0.15\linewidth, edge=output-black,
            [Location Forecasting, middle, text width=0.16\linewidth, edge=output-black, 
                [{DeepMove \cite{feng2018deepmove}, VANext \cite{gao2019predicting}, \cite{song2016deeptransport}, \\Flashback \cite{yang2020location}, MobTCast \cite{xue2021mobtcast}}, leaf, text width=0.37\linewidth, edge=output-black]
            ],
            [Traffic Forecasting, middle, text width=0.16\linewidth, edge=output-black, 
                [{ST-ResNet~\cite{zhang2017deep}, DMVST-Net~\cite{yao2018deep}, STRCN~\cite{jin2018spatio}, \\Periodic-CRN~\cite{zonoozi2018periodic}, \cite{jiang2021deepcrowd}, \cite{liang2019urbanfm}, \cite{yao2019revisiting}, \cite{jiang2019deepurbanevent}, \cite{zhang2023promptst}}, leaf, text width=0.37\linewidth, edge=output-black]
            ]
        ],
        [{Trajectory-related\\Recommendation\\(\secref{subsec:recommendation})}, middle, text width=0.15\linewidth, edge=output-black,
            [Travel Recommendation, middle, text width=0.16\linewidth, edge=output-black,
                [{HRNR~\cite{wu2020learning}, LDFeRR~\cite{liu2021ldferr}, GraphTrip~\cite{gao2023dual}, \cite{zhang2018walking}, \cite{ji2020spatio}}, leaf, text width=0.37\linewidth, edge=output-black]
            ],  
            [Friend Recommendation, middle, text width=0.16\linewidth, edge=output-black,
                [{LBSN2Vec~\cite{yang2019revisiting}, LBSN2Vec++~\cite{yang2020lbsn2vec++}, MVMN~\cite{zhang2020social}, \\TSCI~\cite{gao2018trajectory}, SRINet~\cite{qin2023graph}, FDPL~\cite{rafailidis2018friend}}, leaf, text width=0.37\linewidth, edge=output-black]
            ]
        ],
        [{Trajectory Classification\\(\secref{subsec:classification})}, middle, text width=0.15\linewidth, edge=output-black,
            [Travel Mode Identification, middle, text width=0.16\linewidth, edge=output-black,
                [{TrajectoryNet~\cite{jiang2017trajectorynet}, ST-GRU~\cite{liu2019spatio}, \\TrajODE~\cite{liang2021modeling}, TraClets~\cite{kontopoulos2022traclets}, TrajFormer~\cite{liang2022trajformer}}, leaf, text width=0.37\linewidth, edge=output-black]
            ],
            [Trajectory-User Linking, middle, text width=0.16\linewidth, edge=output-black,
                [{TULER~\cite{gao2017identifying}, TULVAE~\cite{zhou2018trajectory}, DeepTUL~\cite{miao2020trajectory}, \\MainTUL~\cite{chen2022MainTUL}, AdattTUL~\cite{gao2020adversarial}}, leaf, text width=0.37\linewidth, edge=output-black]
            ],
            [Other Perspectives, middle, text width=0.16\linewidth, edge=output-black,
                [{SECA~\cite{dabiri2019semi}, SSFL~\cite{zhu2021semi}, DeepCAE~\cite{markos2020unsupervised}, \\S2TUL~\cite{deng2023s2tul}, DPLink~\cite{feng2019dplink}, AttnTUL~\cite{chen2023trajectory}}, leaf, text width=0.37\linewidth, edge=output-black]
            ]
        ],
        [{Travel Time Estimation\\(\secref{subsec:estimation})}, middle, text width=0.15\linewidth, edge=output-black,
            [Trajectory-based, middle, text width=0.16\linewidth, edge=output-black,
                [{DeepTTE \cite{wang2018will}, DeepTravel \cite{zhang2018deeptravel}, MURAT \cite{li2018multi}, \\CARTE \cite{huang2022context}, TTPNet \cite{shen2020ttpnet}}, leaf, text width=0.37\linewidth, edge=output-black]
            ],
            [Road-based, middle, text width=0.16\linewidth, edge=output-black,
                [{WDR \cite{wang2018learning}, CodriverETA \cite{sun2020codriver},  DeepIST \cite{fu2019deepist}, \\ HetETA \cite{hong2020heteta}, ConSTGAT \cite{fang2020constgat}, CompactETA \cite{fu2020compacteta}}, leaf, text width=0.37\linewidth, edge=output-black]
            ],
            [Other Perspectives, middle, text width=0.16\linewidth, edge=output-black,
                [{ER-TTE~\cite{fang2021ssml}, CatETA \cite{ye2022cateta}, PP-TPU \cite{liu2023uncertainty}}, leaf, text width=0.37\linewidth, edge=output-black]
            ]
        ],
        [{Anomaly Detection \\ (\secref{subsec:detection})}, middle, text width=0.15\linewidth, edge=output-black
            [Offline Detection, middle, text width=0.16\linewidth, edge=output-black,
                [{ATD-RNN~\cite{trajanom_rnn}, IGMM-GAN~\cite{trajanom_gan}, TripSafe\cite{trajanom_tripsafe}}, leaf, text width=0.37\linewidth, edge=output-black]
            ],
            [Online Detection, middle, text width=0.16\linewidth, edge=output-black,
                [{DB-TOD~\cite{trajanom_dbtod}, RL4OASD~\cite{trajanom_subtraj}, \\GM-VSAE~\cite{trajanom_gmvsae}, DeepTEA\cite{trajanom_deeptea}}, leaf, text width=0.37\linewidth, edge=output-black]
            ]
        ],
        [{Mobility Generation\\ (\secref{subsec:generation})}, middle, text width=0.15\linewidth, edge=output-black,
            [Macro-dynamic, middle, text width=0.16\linewidth, edge=output-black,
                [{\cite{simini2021deep}, \cite{chen2020citywide}, \cite{wu2020spatiotemporal}, \cite{zhang2019trafficgan}, \cite{chen2019traffic}, \cite{yin2018gans}}, leaf, text width=0.37\linewidth, edge=output-black]
            ],
            [Micro-dynamic, middle, text width=0.16\linewidth, edge=output-black,
                [{TSG \cite{wang2021large}, DeltaGAN \cite{xu2021simulating}, TrajGen \cite{cao2021generating},COLA~\cite{wang2024cola}, \\DiffTraj \cite{zhu2023difftraj}, LLM-Mob \cite{wang2023would}, LLM-MPE \cite{liang2023exploring}}, middle, leaf, text width=0.37\linewidth, edge=output-black]
            ]
        ]
    ],
  ]
\end{forest}
}
\caption{Taxonomy of this survey with representative trajectory data management and mining works.}
\label{fig:Taxonomy}
\vspace{-4mm}
\end{figure*}

\vspace{-2mm}
\subsection{From Trajectory to Other Formats}\label{sec:data_formart}
The raw trajectory data can be adaptably formatted for various neural network architectures, enhancing its utility in diverse downstream tasks.

\begin{Def}[Matrix]
    For a given city, we can divide it into multiple ($N_1 \times N_2$) grids according to the latitude and longitude. Each grid represents a distinct region within the city.
    Thus, a trajectory can be represented as a continuous sequence of grid identifiers.
    For the origin, destination, and departure time of trajectories, we can construct the Origin-Destination (OD) matrix $\mathcal{M} \in \mathbb{R}^{N_1 \times N_2}$ for any time, where each element represents the inflow and outflow in a particular grid.
\end{Def}

\begin{Def}[Graph]
    A road network for a city can be converted into a directed graph of roads $\mathcal{G} = (\mathcal{V}, \mathcal{A})$, where $\mathcal{V}$ denotes the roads in the network, and $\mathcal{A}$ represents the connectivity between the road segments.
    Consequently, $\mathcal{A}_{ij} = 1$ if and only if road $i$ and $j$ can be directly connected.
    In this setting, the trajectory can be extracted as a sequence of roads based on the road segments that the trajectory passes through.
\end{Def}

\begin{Def}[Raster]
    A raster image, denoted as $\mathcal{I} \in \mathbb{R}^{H\times W \times C}$, is composed of pixels arranged in a grid. Each pixel possesses specific semantic and positional information, forming the entire image in a predetermined order. Thus, trajectories can naturally be transformed into raster images. A simple and intuitive approach involves treating the entire map as a binary image, where pixels traversed by the trajectory are set to 1, and those not traversed are set to 0. Effective rasterization primarily considers trajectory shape, speed, and direction, which has been extensively studied in the literature~\cite{endo2016classifying}.
\end{Def}


Trajectories can also be represented in other vision forms, such as converting them into bird's-eye view maps. However, this type of data is more closely related to computer vision and receives less attention in the trajectory data mining and management community. Therefore, we do not include this type of purely visual form here.

\vspace{-1mm}
\section{Overview and Categorization}~\label{sec:tax}
\vspace{-3mm}

The taxonomy of representative works this survey paper is presented in \figref{fig:Taxonomy}. Furthermore, we summarize the carefully designed structured content of this paper:

\begin{itemize}[leftmargin=*]
\item \textbf{Deep Learning for Trajectory Data Management.} Deep learning is seamlessly integrated into all phases of trajectory management—including \textit{pre-processing}, efficient \textit{storage}, high-quality \textit{analytics}, and clear \textit{visualization}—to facilitate subsequent mining tasks.
\item \textbf{Deep Learning for Trajectory Data Mining.} Integrating deep learning enables comprehensive solutions for six major trajectory mining tasks: \textit{forecasting}, \textit{recommendation}, \textit{classification}, \textit{travel time estimation}, \textit{anomaly detection}, and \textit{mobility generation}.
\item \textbf{Advances in Large Models for Trajectory Computing.} The emergence of FMs and LLMs has similarly transformed the trajectory community, giving rise to new research questions and technologies that we systematically summarize to provide a cutting-edge perspective.
\item \textbf{Applications \& Resources.} Deep learning interlinks trajectory computing to generate practical applications in diverse domains, such as \textit{personal services}, \textit{business platforms}, and \textit{policy guidance}. We also provide a comprehensive exploration of publicly available \textit{datasets} and \textit{tools}.
\end{itemize}



\section{Deep Learning For Trajectory Data Management}\label{sec:management}

\subsection{Pre-Processing}\label{subsec:pre-processing}

Recorded trajectories aim to depict the actual movements of objects. However, inherent inaccuracies arise from sampling devices and environmental uncertainties. Pre-processing refines raw data by simplifying redundant and anomalous points, completing missing ones, and employing map matching for calibration, meeting specific needs.

\begin{figure}[t!]
 \centering
 \includegraphics[width=0.85\linewidth]{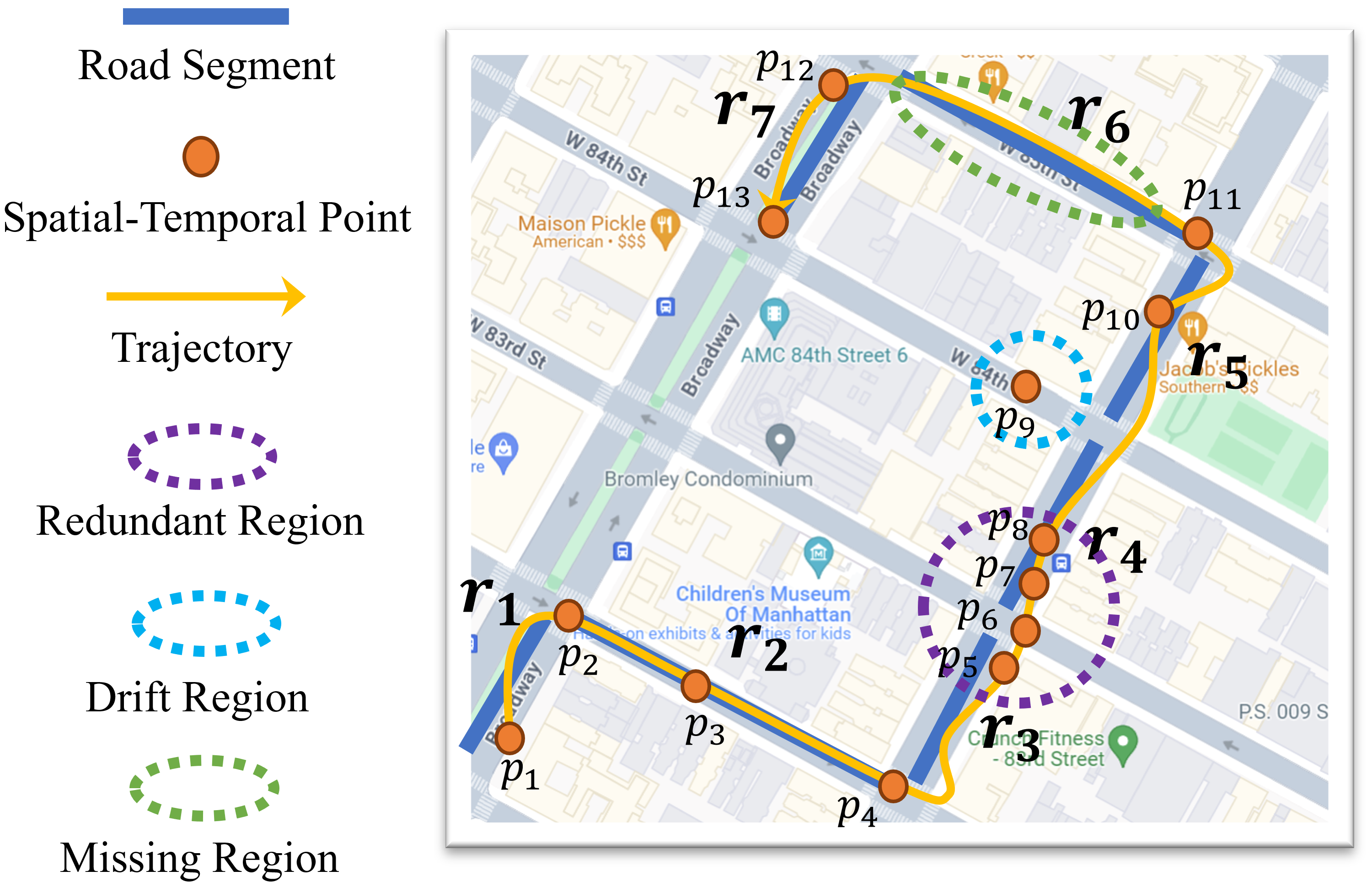}
 \caption{Pre-Processing example.}
 \label{fig:pre-process}
 \vspace{-6mm}
\end{figure}

\textbf{Trajectory Simplification.} In the presence of sensor noise and the inherent characteristics of high-frequency sampling, Fig~\ref{fig:pre-process} illustrates the emergence of nearly identical "redundant points" (\eg, $p5-p8$) and "drift points" (\eg, $p9$) within a moving trajectory. To mitigate these issues, \textit{trajectory simplification methods are designed to remove redundant and anomalous points, effectively reducing data without significantly altering the overall information of trajectory}.

Early methods relied on human-crafted rules, divided into batch mode (accessing complete data to balance compression and loss) and online mode (accessing only a buffer for real-time compression). Notable batch methods include DP~\cite{trajsimp_dp} and DPTS~\cite{long2013direction}, which compute point importance, while online techniques use sliding windows~\cite{keogh2001online} and normal opening windowing~\cite{meratnia2004spatiotemporal} algorithms to extract feature points. Semantic simplification~\cite{li2021trace} offers an alternative by leveraging road networks to reduce spatial redundancy. To overcome the lack of adaptability in rule-based methods, recent studies utilize deep learning, such as RLTS~\cite{trajsimp_zhengwang_rl} and S3~\cite{trajsimp_lightweight}, to minimize the error between original and simplified trajectories under length constraints. Furthermore, MARL4TS~\cite{trajsimp_errorbound} minimizes simplified trajectory length under bounded error conditions, and RL4QDTS~\cite{wang2023collectively} introduces query accuracy-driven trajectory simplification using multi-agent reinforcement learning to tackle storage costs and expedite query processing.

\textbf{Trajectory Recovery.} Due to issues with recording devices such as communication latency, GPS localization errors, and privacy issues, the collected data usually covers a substantial number of trajectories with low or missing sample rates \cite{wang2021survey}. Take Fig~\ref{fig:pre-process} as an example, the raw trajectory lacks any recorded information within the green dashed region (\eg, driving in areas with missing signal stations), which may hinder its utilization for downstream applications. To this end, \textit{trajectory recovery aims to transform these irregular, low-sampled trajectories into high-sampled ones, effectively supporting mobility computing applications.}

Trajectory recovery \cite{long2016kinematic}, traditionally viewed as spatial series data completion, relies on correlations between adjacent points to impute missing values. Early methods, such as linear and polynomial~\cite{tremblay2006interpolation} interpolations, were limited in capturing complex dependencies. Recent deep models have advanced sparse trajectory completion. Trajectory recovery is typically categorized based on external information. The first category, free-space trajectory recovery, focuses on modeling intricate transition patterns within trajectory sequences. DHTR~\cite{wang2019deep} extended the Seq2Seq framework to Sub-Seq2Seq, employing a deep hybrid model with a Kalman filter for uncertainty reduction. To address sparsity, AttnMove~\cite{xia2021attnmove} proposed an attention-based model integrating historical and periodic patterns, using Bayesian neural networks for uncertainty estimation. PeriodicMove~\cite{sun2021periodicmove} introduced a GNN-based attention model that learns complex location transitions from directed graphs constructed from trajectories. TrajBERTT~\cite{si2023trajbert} and TEIR~\cite{chen2023teri} leverage Transformer architectures to refine spatio-temporal modeling, applicable even without explicit geographical coordinates or with variable sampling rates.

The second setting, map-constrained recovery, involves utilizing external knowledge, such as road networks, to map segments or points of interest. MTrajRec~\cite{ren2021mtrajrec} pioneered multi-task learning within Seq2Seq models for this setting, incorporating modules for constraint masking, attention, and attribute enhancement. RNTrajRec~\cite{chen2023rntrajrec} further introduced a novel spatio-temporal transformer network, GPSFormer, seamlessly integrated with a new road network representation model, GridGNN. Additionally, significant semantic and visual information can enhance recovery. STR~\cite{long2024learning} and VisionTraj~\cite{li2023visiontraj} address this using a heterogeneous information network encoder to model semantic correlations. Beyond this, Traj2Traj~\cite{liao2023traj2traj} utilizes a latent factor module to improve recovery efficiency, and PATR~\cite{zhang2022patr} incorporates a periodic perception module for real logistics platforms.

An important related application is the recovery of urban road networks. DeepMG \cite{ruan2020learning} is a representative approach that discovers and extracts the underlying road network structure from extensive trajectory data. Furthermore, studies like DF-DRUNet~\cite{li2024df} and DelvMap~\cite{wang2024delvmap} utilize deep neural networks for multimodal fusion of satellite images and trajectory data to improve road network recovery performance.

\textbf{Map-Matching,} \textit{which converts spatio-temporal points' latitude and longitude sequences into road segment sequences,} facilitating downstream intelligent transportation tasks.  As illustrated in Fig~\ref{fig:pre-process}, the original trajectory sequence $\{p_1, ..., p_{13}\}$ can be mapped to road segments $\{r_1, ..., r_{7}\}$. 

Most prior studies on map matching progressed from geometric~\cite{taylor2001road} and topological~\cite{quddus2003general} approaches to probabilistic statistical algorithms~\cite{ochieng2003map}. Hidden Markov Models (HMMs), specifically, demonstrate superior robustness to noise and varying sampling rates~\cite{mm-fmm}. However, HMM-based methods do not fully utilize abundant trajectory data. DeepMM~\cite{mm-deepmm} introduced the first deep model using an attention mechanism for accurately mapping sparse and noisy trajectories onto the road network. Addressing the scarcity of well-matched data, a Transformer-based model~\cite{mm-transformer} employs transfer learning, pre-training on generated data and fine-tuning with limited labeled samples. Besides, L2MM~\cite{jiang2023l2mm} proposes high-frequency and data distribution augmentation to improve the model's generalization for map matching. Nevertheless, these methods overlook the graph nature of the problem. GraphMM~\cite{mm-graph} incorporates graph neural networks to extract intra-trajectory, inter-trajectory, and trajectory-road correlations. Beyond this, DMM~\cite{mm-dmm}, TBMM~\cite{zhu2023map}, and FL-AMM~\cite{mm-fl} extend map matching to scenarios involving wireless sensor data by integrating techniques like federated and reinforcement learning.

\subsection{Storage}\label{subsec:storage}

To cope with the surge in streaming trajectory data, research in trajectory storage, indexing, and querying  remains crucial.

\textbf{Storage Database.} Traditional trajectory storage systems focus on the spatio-temporal point level, leading to numerous systems designed for storing and querying trajectory data. The research community has developed specialized management systems~\cite{wang2014sharkdb,ding2018ultraman,fang2021dragoon} for specific trajectory data types, although the supported query types are often limited. Concurrently, the open-source community has extended existing distributed systems~\cite{li2021trajmesa} for large-scale trajectory storage by introducing custom data formats like LineString and GPX~\cite{foster2004gpx}.

Vector databases, capitalizing on deep representation learning, have become a prevalent database type~\cite{wang2021milvus}, offering efficient storage, retrieval, and querying capabilities for diverse data. Limited research has focused on trajectory vector databases~\cite{cai2017vector,fang2023ghost}, with current efforts primarily directed at advancing trajectory representation learning to automatically compress raw trajectories into low-dimensional vector spaces. Since trajectory vector quality is typically assessed by similarity, further details are elaborated in Section~\ref{similarity}.

\textbf{Index \& Query.}
\textit{Trajectory indices are data structures designed to efficiently organize and store trajectories, enabling quick retrieval and analysis}. They are vital for optimizing the search performance of various trajectory queries, including similarity search, $k$-nearest neighbor query, and similarity join.

While TraSS~\cite{he2022trass} recently introduced a novel spatial index XZ for rapid trajectory querying in a key-value database, conventional trajectory indices~\cite{chen2015efficient,xie2017distributed,chen2018price,yuan2019distributed} primarily adapt R-trees to hierarchically organize trajectory points or segments, thereby accelerating queries by narrowing search areas. Consequently, deep learning has been extensively applied to enhance data indices concerning query efficiency, resulting in learned indices that model data distribution and access patterns. Existing spatial learned indices~\cite{qi2020effectively,pandey2020case} predominantly focus on low-dimensional data, such as two-dimensional GPS points. X-FIST~\cite{ramadhan2022x} extended the learned index concept to trajectories by indexing their Minimum Bounding Rectangles (MBRs). For each trajectory, X-FIST first generates a list of sub-trajectories, then constructs two Flood indices on the lower-left and upper-right vertices of the sub-trajectory MBRs.

\subsection{Analytics}\label{subsec:analytics}

Efficient and precise similarity measurement, as well as clustering analysis, are foundational for various mining tasks involving complex and multi-source trajectory data.

\begin{table}[b!]
\vspace{-2mm}
\captionsetup{font=small}
\caption{Classification of existing trajectory similarity measures.
$m$ and $n$ denote the numbers of points in two trajectories, respectively. $i_m$ and $i_n$ denote image sizes. $k_m$ and $k_n$ denote the numbers of neighbor nodes on the road network graph. Note that, the dimensionality of trajectory embeddings is a small constant and thus it does not affect time complexity results.
}

\label{tab:similarity_compare}
\vspace{-2mm}
\setlength{\tabcolsep}{0.5mm}{}
\renewcommand\arraystretch{1.3}
\begin{small}
\scalebox{0.72}{
\centering
\begin{tabular}{c|c|c|cccc}
\toprule

\multicolumn{3}{c|}{\textbf{Category}} & \textbf{Method} & \textbf{Complexity} & \textbf{Robustness} & \textbf{Components}\\

\midrule

\multirow{5}{*}{\rotatebox{90}{Heuristic}}
& \multicolumn{2}{c|}{\multirow{3}{*}{Point-based}}
& DTW~\cite{chang2024trajectorysimm} & $O(mn)$ & \rxmark & - \\
& \multicolumn{2}{c|}{~}
& LCSS~\cite{chang2024trajectorysimm} & $O(mn)$ & \gcmark & - \\
& \multicolumn{2}{c|}{~}
& EDR~\cite{chang2024trajectorysimm} & $O(mn)$ & \gcmark & - \\
\cline{2-7}
& \multicolumn{2}{c|}{\multirow{2}{*}{Shape-based}}
& Fréchet~\cite{chang2024trajectorysimm} & $O(mn)$ & \rxmark & - \\
& \multicolumn{2}{c|}{~}
& Hausdorff~\cite{chang2024trajectorysimm} & $O(mn)$ & \rxmark & - \\

\midrule

\multirow{20}{*}{\rotatebox{90}{Learning}}
& \multirow{14}{*}{\rotatebox{90}{Free Space}}
& \multirow{9}{*}{SSL-based}
& t2vec~\cite{trajsimi_t2vec} & $O(m+n)$ & \gcmark & RNNs \\
& & & RSTS~\cite{trajsimi_rsts} & $O(m+n)$ & \gcmark & RNNs \\
& & & At2vec~\cite{trajsimi_at2vec} & $O(m+n)$ & \gcmark & RNNs \\
& & & Play2vec~\cite{trajsimi_play2vec} & $O(m+n)$ & \gcmark & RNNs \\
& & & CL-Tsim~\cite{trajsimi_cltsim} & $O(m+n)$ & \gcmark & RNNs \\
& & & TrjSR~\cite{trajsimi_trjsr} & $O(i_m+i_n)$ & \gcmark & CNNs \\
& & & CSTRM~\cite{trajsimi_cstrm} & $O(m^2+n^2)$ & \gcmark & Attention \\
& & & TrajCL~\cite{trajsimi_trajcl} & $O(m^2+n^2)$ & \gcmark & Attention \\
& & & TrajRCL~\cite{li2023self} & $O(m^2+n^2)$ & \gcmark & Attention \\
\cline{3-7}
& & \multirow{5}{*}{SL-based}
& NEUTRAJ~\cite{trajsimi_neutraj} & $O(m+n)$ & \gcmark & RNNs\\
& & & Traj2SimVec~\cite{trajsimi_traj2simvec} & $O(m+n)$ & \gcmark & RNNs\\
& & & TMN~\cite{trajsimi_tmn} & $O(mn)$ & \gcmark & RNNs\\
& & & T3S~\cite{trajsimi_t3s} & $O(m^2+n^2)$ & \gcmark & Attn.+RNNs \\
& & & TrajGAT~\cite{trajsimi_trajgat} & $O(mk_m+nk_n)$ & \gcmark & GNNs\\
\cline{2-7}
& \multirow{7}{*}{\rotatebox{90}{Road Network}}
& \multirow{2}{*}{SSL-based}
& Trembr~\cite{trajsimi_trembr} & $O(m+n)$ & \gcmark & RNNs \\
& & & LightPath~\cite{yang2023lightpath} & $O(m^2+n^2)$ & \gcmark & Attention \\
\cline{3-7}
& & \multirow{5}{*}{SL-based}
& GTS~\cite{trajsimi_gts} & $O(m+n)$ & \rxmark & GNNs+RNNs \\
& & & GTS+~\cite{trajsimi_gts} & $O(m+n)$ & \rxmark & GNNs+RNNs \\
& & & GRLSTM~\cite{trajsimi_grlstm} & $O(m+n)$ & \rxmark & GNNs+RNNs \\
& & & SARN~\cite{trajsimi_sarn} & $O(m+n)$ & \rxmark & GNNs+RNNs \\
& & & ST2Vec~\cite{trajsimi_st2vec} & $O(m^2+n^2)$ & \rxmark & GNNs+RNNs+Attn. \\

\bottomrule
\end{tabular}

}
\end{small}

\end{table}

\begin{figure}[t!]
 \vspace{-4mm}
 \centering
 \includegraphics[width=1.0\linewidth]{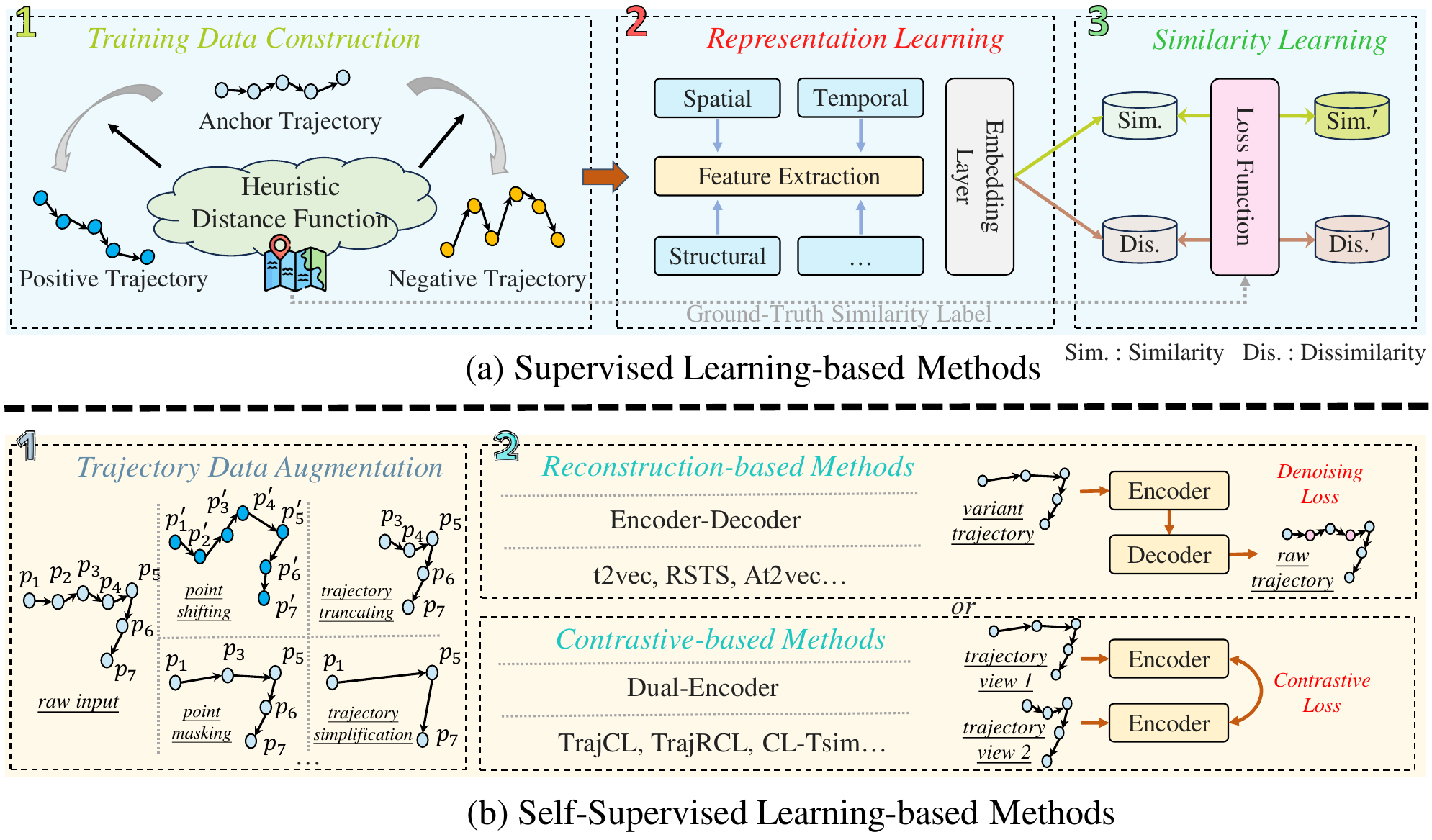}
 \caption{Different pipeline of trajectory similarity methods.}
 \label{fig:similarity}
 \vspace{-4mm}
\end{figure}

\textbf{Similarity Measurement.}\label{similarity}
\textit{Trajectory similarity quantifies trajectory resemblance using a set of distance metrics.} Traditional heuristic methods~\cite{chang2024trajectorysimm} include point-based (DTW, LCSS, EDR) and shape-based (Fréchet, Hausdorff) distances. Recently, deep learning studies~\cite{trajsimi_t2vec,trajsimi_rsts,trajsimi_at2vec} have enhanced measurement effectiveness and computational efficiency. As shown in Figure~\ref{fig:similarity}, these methods are classified by learning paradigm (Self-Supervised Learning, SSL; or Supervised Learning, SL) and metric space (Free Space or Road Network). We combine these perspectives to detail specific measures for each category, summarizing differences in Table~\ref{tab:similarity_compare}.

$\circ$ \emph{Free Space}: These methods measure the similarity of raw trajectories in free space, often converting them to cell trajectories for processing. They are categorized into SSL-based and SL-based approaches. SSL-based methods learn robust trajectory representations directly from unlabeled trajectories without relying on heuristic rules. t2vec~\cite{trajsimi_t2vec} pioneered the adoption of SSL by generating similar training pairs through subsampling. RSTS~\cite{trajsimi_rsts}, Tedj~\cite{trajsimi_tedjopurnomo}, and At2vec~\cite{trajsimi_at2vec} enhanced t2vec by incorporating time, multi-granularity, and POI similarity, respectively; these are generally considered reconstruction-based. Recently, CL-Tsim~\cite{trajsimi_cltsim} introduced a contrastive method to learn discriminative representations by creating positive and negative training samples. TrajCL~\cite{trajsimi_trajcl} and TrajRCL~\cite{li2023self} introduced various augmentation and contrastive learning methods to jointly capture spatial and structural similarity. Furthermore, CSTRM~\cite{trajsimi_cstrm} used a shallow transformer encoder, while TrjSR~\cite{trajsimi_trjsr} transformed trajectories into images; both aim to capture multi-scale similarity. For practical scenarios, Play2vec~\cite{trajsimi_play2vec} learns motion trajectory similarity, applying it to sports match analysis. SL-based methods efficiently approximate existing heuristic measurements. NEUTRAJ~\cite{trajsimi_neutraj}, the pioneer, adapted an RNN with a spatial attention memory module to learn correlations between spatially proximate trajectories. Traj2SimVec~\cite{trajsimi_traj2simvec} and~\cite{trajsimi_chen_iotj} improved NEUTRAJ's pre-processing and training efficiency. Subsequently, TMN~\cite{trajsimi_tmn} proposed an attention-based matching module to directly learn point-to-point correlation between two trajectories. T3S~\cite{trajsimi_t3s} combined LSTM with self-attention to learn representations in Euclidean and grid spaces. Unlike prior methods, TrajGAT~\cite{trajsimi_trajgat} introduced a graph-based self-attention model, representing trajectories as graphs, and used a quadtree to partition space for capturing fine-grained dependencies.

$\circ$ \emph{Road Networks}: These methods measure similarity for trajectories mapped onto road networks, primarily suitable for urban individual or vehicle movements. They are also categorized by learning paradigms. SSL-based methods include Trembr~\cite{trajsimi_trembr} and LightPath~\cite{yang2023lightpath}, utilizing RNN and Transformer-based Seq2Seq models, respectively. Constrained by the underlying road network, they encode intrinsic spatial and temporal properties into a latent space. Following this framework, PIM~\cite{yang2021unsupervised} reduced reliance on vast training data via a course-wise negative sampling strategy. WSCCL~\cite{yang2022weakly} incorporated a weakly supervised contrastive learning model to train a temporal path encoder, addressing label acquisition difficulty. For SL-based methods, GTS~\cite{trajsimi_gts} pioneered by defining various road network trajectory similarities, then employing GCN and LSTM to learn embeddings for POI sequences in the graph. Building on this, GRLSTM~\cite{trajsimi_grlstm} and GTS+~\cite{zhou2023spatial} utilize knowledge graphs and spatio-temporal LSTM with time gates, respectively, to jointly capture trajectory and road network attributes. ST2Vec~\cite{trajsimi_st2vec} is another study focusing on spatio-temporal similarity on road networks, differing from GTS+ by integrating spatial and temporal features before LSTM input. Furthermore, SARN~\cite{trajsimi_sarn} focuses on learning segment embeddings, proposing a contrastive learning-based GCN to capture local and global road segment similarities.

\begin{figure}[htbp!]
 \vspace{-2mm}
 \centering
 \includegraphics[width=0.9\linewidth]{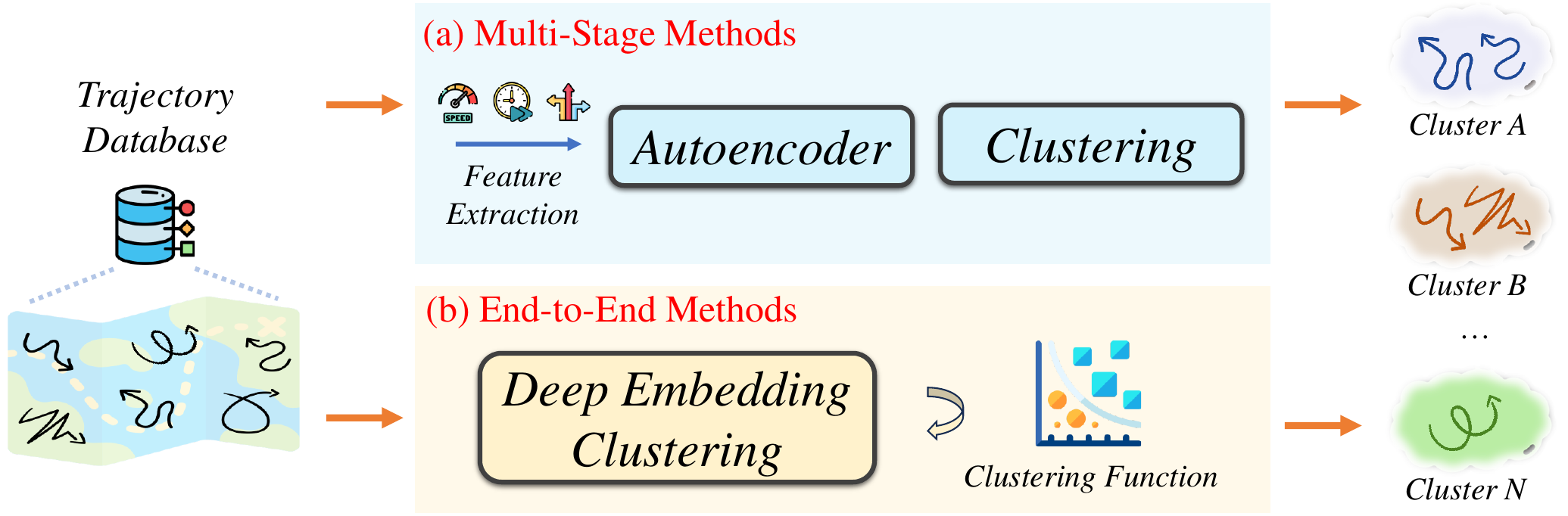}
 \caption{Different pipeline of cluster analysis methods.}
 \label{fig:cluster}
 \vspace{-2mm}
\end{figure}

\textbf{Cluster Analysis.}
\textit{Trajectory clustering groups trajectories based on their similarity, ensuring high intra-cluster similarity}~\cite{yuan2017review}. Traditional methods rely heavily on the choice of similarity measurement, which often leads to variable clustering quality. Current learning-based methods are robust to spatio-temporal scale variations by accounting for latent trajectory features. As depicted in Figure~\ref{fig:cluster}, these methods are generally categorized into multi-stage and end-to-end approaches based on their processing workflows.

\par In general, multi-stage trajectory clustering methods~\cite{trajclus_trip2vec} typically involve two steps: first, extracting low-dimensional representations by using a sliding window to capture robust movement features, which are then fed into an LSTM-based Autoencoder (AE) to learn fixed-length representations. Subsequently, a traditional clustering algorithm, such as K-means~\cite{arthur2007k}, is applied to the learned representations. Trip2Vec~\cite{trajclus_trip2vec} extracts three trip attributes (time, origins, destinations), inputs them into a fully connected AE to generate trajectory representations, and then uses K-means for clustering. In contrast, end-to-end approaches directly integrate Deep Embedding Clustering (DEC) to simultaneously refine trajectory representations and clustering assignments. For instance, the study~\cite{trajclus_olive2020deep} applies AE-based t-SNE and DEC for aircraft trajectory clustering. DETECT~\cite{trajclus_DETECT} uses an LSTM-based AE and environmental context to jointly refine embeddings and clustering. E2DTC~\cite{trajclus_E2DTC}, an RNN-based AE method, introduces a dedicated triplet loss for clustering.

\subsection{Visualization}\label{subsec:visualization}

To enable real-time visualization and interactive analysis of extensive mobility data, traditional trajectory visualization methods rely on temporal and spatial dimensions of geo-points. Techniques like density maps, heatmaps, and spatio-temporal cubes~\cite{bach2017descriptive} are used for macroscopic analysis.

However, visualizing large raw trajectories can result in information redundancy, as addressed by deep clustering and simplification methods in Sec~\ref{subsec:pre-processing} and~\ref{subsec:analytics}. Utilizing these methods, researchers can obtain grouped trajectories, allowing for a detailed examination of various trajectory movement patterns. For instance, ~\cite{lee2019visual} presents an interactive system called Surveillance that uses LSTM models and network embedding to detect and visualize urban congestion conditions. Similarly,~\cite{zhou2018visual} uses an iterative sampling scheme for OD flows, creating meaningful visual encodings. Deep learning's capability to extract hidden knowledge from data without prior knowledge is also leveraged to assist in individual trajectory visual exploration.~\cite{maekawa2020deep} introduces DeepHL, employing attention-based neural networks for automatic detection and visualization of meaningful trajectory segments. In DSAE~\cite{liu2017visualization}, a deep sparse autoencoder extracts hidden features, mapping them to the RGB color space to visualize driving behavior. Later,~\cite{zhang2021deep} uses GIS map integration to enhance anomaly visualization. More analyses of trajectory visualization are discussed in~\cite{deng2023survey}.

\section{Deep Learning for Trajectory Data Mining}\label{sec:mining}

\subsection{Trajectory-related Forecasting}\label{subsec:forecasting}

\begin{figure}[htbp!]
 \centering
 \vspace{-2mm}
 \includegraphics[width=0.9\linewidth]{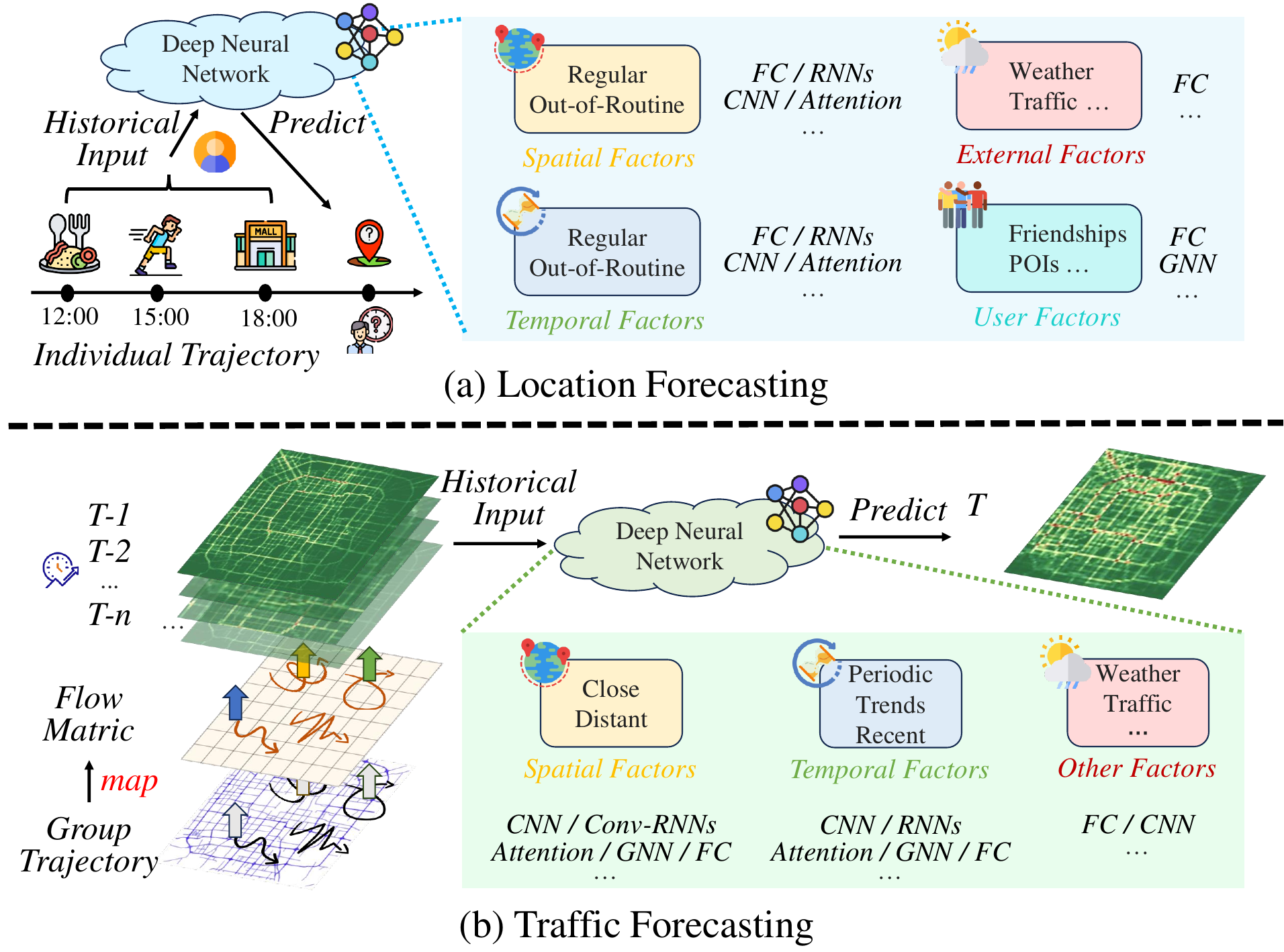}
 \caption{Forecasting tasks schematic and influencing factors.}
 \label{fig:prediction}
 \vspace{-3mm}
\end{figure}

As shown in Figure~\ref{fig:prediction}, in trajectory data mining, \textit{forecasting tasks aim to accurately predict future movements of individuals (i.e., Location Forecasting) or crowds (Flow Forecasting) based on historical data}\cite{luca2021survey}.

\textbf{Location Forecasting.} This task aims to predict an individual's subsequent location using their historical movement data. It requires modeling spatial (\eg, location), temporal (\eg, day of week), external (\eg, weather), and personalized (\eg, periodic visits) patterns. \textit{Formally, given a person's historical movement data, the task predicts the most likely next spatial point or region}~\cite{luca2021survey}. This is formulated either as a classification problem (next location is one of predefined regions) or a regression problem (predicting exact geographical coordinates). The challenge lies in accurately modeling the complexity of human movement and factors influencing location selection, such as time, personal preferences, and social behavior.

Intuitively, classification models learn from a historical sequence of visited locations, integrating various learning blocks (CNN \cite{bao2021bilstm}, RNN \cite{yang2020location}, ST-RNN \cite{liu2016predicting}, Embedding \cite{yang2020location}, Attention mechanisms \cite{feng2018deepmove}, and GNNs~\cite{limcn4rec}) to capture the transfer probability distribution of all possible locations; the highest probability indicates the most likely next visit. For instance, DeepMove~\cite{feng2018deepmove}, an attentional recurrent neural network, uses two attention mechanisms to capture multi-level periodicity and utilizes GRUs for trajectory processing and prediction. Flashback~\cite{yang2020location}, a general RNN architecture, addresses user movement sparsity by using spatio-temporal context within the RNN to identify hidden states with high predictive power. Unlike classification-based methods, regression-based approaches forecast continuous and exact values representing the next spatial point. For example, Song \textit{et al.}~\cite{song2016deeptransport} introduced a multi-task deep learning framework with stacked LSTM layers to simultaneously predict future traffic patterns and positions. Considering the influence of multiple contexts, MobTCast~\cite{xue2021mobtcast} uses the transformer architecture as a spatio-temporal feature extractor to process both temporal and semantic contexts.

Additionally, recent studies have extended location forecasting into two variants: next POI recommendation and incomplete path prediction. The former primarily addresses cold start scenarios and user preference recommendations. The latter is applied in contexts like food delivery and logistics, predicting a series of locations based on a worker's current incomplete route tasks, which adds complexity. For detailed discussion, refer to~\cite{islam2022survey} and~\cite{wen2023survey}.

\textbf{Traffic Forecasting.}
Traffic forecasting predicts the movement and density of traffic within a given area over time. This task analyzes historical mobility data, congestion patterns, and flow trends to predict the number of entities that will congregate in an area at a future time. \textit{Formally, traffic forecasting is typically treated as a time series forecasting problem, aiming to predict future flows based on past observations}~\cite{chen2025stttc,chen2025select}. The complexity arises from the dynamic and unpredictable nature of traffic flow, influenced by factors such as time of day, weather, accidents, and construction.

In practice, trajectories are first transformed into matrices based on time and corresponding regions (as outlined in Sec. \ref{sec:data_formart}). Forecasts can then be made using classical time series models like Autoregressive Moving Average and Vector Autoregressive~\cite{canova1999vector}. However, these methods struggle with spatial dependencies and various additional features (\eg, weather), limiting their performance. Since traffic flows are matrix-formatted, CNNs can effectively capture their local and global spatio-temporal dependencies. Additionally, RNN models like LSTM can model complex temporal dynamics. Thus, deep learning methods efficiently capture patterns in the temporal evolution of crowd flows. ST-ResNet \cite{zhang2017deep} pioneered traffic flow prediction using deep neural networks, utilizing residual CNN units to capture temporal closeness, trends, and periodic patterns. The output from each attribute type is aggregated with external factors to predict the flow. Follow-up studies include DMVST-Net~\cite{yao2018deep}, which investigates multi-view spatio-temporal patterns. STRCN~\cite{jin2018spatio} combines CNN and LSTM for spatio-temporal modeling and assigns weights to different branches. Periodic-CRN~\cite{zonoozi2018periodic} focuses on capturing repeated periodic patterns. Furthermore, numerous deep learning-based methods have emerged, broadly categorized as ConvLSTM-based~\cite{jiang2021deepcrowd}, multi-task-based~\cite{jiang2019deepurbanevent}, and attention-based methods~\cite{fang2022transfer}. Recent studies also increasingly utilize spatio-temporal graphs to model traffic flows~\cite{sun2020predicting,fang2021mdtp,chen2025eac}, leveraging the advantages of GNNs.

\subsection{Trajectory-related Recommendation}\label{subsec:recommendation}

\begin{figure}[t!]
 \vspace{-2mm}
 \centering
 \includegraphics[width=0.9 \linewidth]{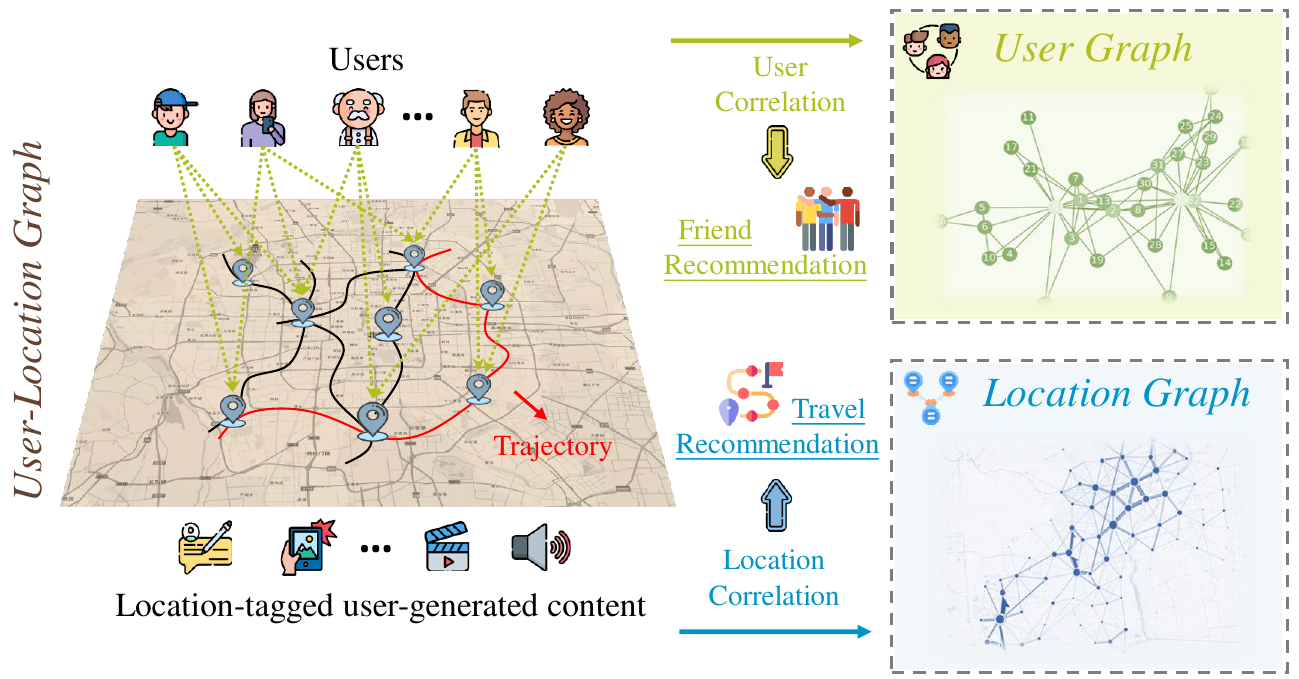}
 \caption{Recommendation of location-based social networks.}
 \label{fig:lbsn}
 \vspace{-5mm}
\end{figure}

As illustrated in Figure~\ref{fig:lbsn}, critical tasks within location-based service systems (LBSN) encompass travel and friend recommendations. \textit{Travel recommendation aims to provide suitable routes based on user constraints and preferences. Friend recommendation infers social relationships and suggests potential acquaintances based on user mobility patterns and behaviors.} Analyzing historical trajectory data, social connections, and potential needs is key to delivering precise recommendations that enhance user travel and social interactions.

\textbf{Travel Recommendation.} \textit{The primary objective of travel recommendation is to generate POI sequences tailored to specific traveler constraints, such as duration, origin, destination, and visitation targets}. Traditionally termed travel query and planning, this domain seeks to maximize user satisfaction by solving the orienteering problem. The core methodology employs heuristics to integrate POIs with trajectories. These approaches generally fall into five categories: search-based, probability-based, biomimetic-based, clustering-based, and constraint-based methods~\cite{zhang2024survey}, many of which originate from robotic pathfinding.

The proliferation of ubiquitous tracking devices has facilitated data-driven, personalized travel recommendations. Early hybrid approaches utilize neural networks to approximate A* search cost functions; notably, HRNR~\cite{wu2020learning} models complex traffic data for efficient route planning. Deep learning has inspired diverse architectures: sequential models~\cite{liu2021ldferr,fu2021progrpgan,wang2021personalized,wang2022personalized,wang2023query2trip} employ RNNs to extract POI features under diversity constraints. For instance, LDFeRR~\cite{liu2021ldferr} combines GRUs with attention mechanisms to optimize fuel efficiency in long-distance travel. Conversely, graph-based methods~\cite{gao2023dual,wu2021learning,wu2019learning} capture spatial dependencies; GraphTrip~\cite{gao2023dual} leverages spatio-temporal graphs and transfer learning to mitigate data sparsity. Multi-modal approaches~\cite{zhang2018walking,liu2020multi} enhance performance by integrating text and imagery, with \cite{zhang2018walking} pioneering the use of Google Street View data. Furthermore, reinforcement learning frameworks~\cite{ji2020spatio,bi2019evacuation} frame urban routing as a decision-making process, as demonstrated by ~\cite{ji2020spatio} adaptive deep reinforcement learning method.

\textbf{Friend Recommendation.} Friend recommendation in location-based social networks (LBSN) enhances user engagement by capitalizing on the correlation between frequent co-visitation and social formation~\cite{bao2015recommendations}. \textit{Formally, literature typically frames this task as social relationship inference, estimating friendship probability based on historical check-in trajectories and existing social networks. Certain studies extend this objective to top-k friend recommendation.}

Early research relied on location co-occurrence or interest similarity. Brown~\textit{et al.}~\cite{brown2012online} established the correlation between geographic proximity and social ties, while Chu~\textit{et al.}~\cite{chu2013friend} incorporated dwell time to analyze location similarity. Yu~\textit{et al.}~\cite{yu2011geo} utilized random walks on heterogeneous information networks merging GPS data to estimate link relevance. With the advent of graph neural networks (GNNs)~\cite{perozzi2014deepwalk,grover2016node2vec,velickovic2017graph}, research has pivoted toward nonlinear representations of LBSN. LBSN2Vec~\cite{yang2019revisiting} and its extension LBSN2Vec++~\cite{yang2020lbsn2vec++} employ hypergraphs to integrate user, temporal, and spatial semantics for automated feature representation. Similarly, MVMN~\cite{zhang2020social} proposes a multi-view matching network integrating diverse factors. Addressing data sparsity, TSCI~\cite{gao2018trajectory} leverages VAE latent variables to estimate friendship trajectory distributions. Recently, SRINet~\cite{qin2023graph} introduced a GNN framework to mitigate noise in mobility data, while FDPL~\cite{rafailidis2018friend} frames recommendation as a ranking task, utilizing deep pairwise learning based on Bayesian personalized ranking.

\subsection{Trajectory Classification}\label{subsec:classification}

\textit{Trajectory classification aims to distinguish trajectory characteristics by learning latent patterns from historical data to categorize new trajectories.} Categories encompass transportation modes, animal types, and specific users~\cite{da2019survey}. Early methodology was predominantly heuristic; for instance, TraClass~\cite{lee2008traclass} utilizes adaptive spatial grids, while subsequent work~\cite{soleymani2014integrating} employs uneven spatio-temporal gridding based on predefined thresholds. Other studies construct recognition models (\eg, SVM, Random Forests) using features such as distance and heading change rates~\cite{zheng2008learning,dodge2009revealing}. To better capture complex spatio-temporal and semantic dependencies, recent deep learning advancements prioritize Travel Mode Identification (TMI) and Trajectory User Linking (TUL).

\textbf{Travel Mode Identification.} TMI focuses on categorizing movement patterns from raw trajectories, accounting for mode transitions within a single journey (\eg, cycling followed by transit). \textit{Formally, given an individual's historical trajectory records, TMI task aims to identify the potential movement modes encompassed in the journey}~\cite{hu2022estimator}. This problem is typically formulated as a multi-class or multi-label classification task. Primary challenges arise from irregular sampling intervals and inherent spatio-temporal noise.

Existing methods achieve high accuracy by employing AE~\cite{dabiri2019semi,liu2023distributional}, RNN~\cite{jiang2017trajectorynet,liu2017end,liu2019spatio,liang2021modeling}, CNN~\cite{kontopoulos2022traclets,zeng2023trajectory}, Attention~\cite{liang2022trajformer,jiang2020multi}, and GNN~\cite{yu2023graph} architectures. Early sequence modeling approaches, such as TrajectoryNet~\cite{jiang2017trajectorynet} and bidirectional LSTM classifiers~\cite{liu2017end}, utilize segment information and data normalization. Addressing the limitations of discrete-time updates, ST-GRU~\cite{liu2019spatio} incorporates segment-wise gating, while TrajODE~\cite{liang2021modeling} leverages neural ordinary differential equations to model continuous temporal dynamics. Alternatively, TraClets~\cite{kontopoulos2022traclets} converts trajectories into raster images for CNN-based classification. Recently, TrajFormer~\cite{liang2022trajformer} adapts the transformer architecture with squashing functions to balance efficiency and accuracy.

\begin{table}[t!]
\captionsetup{font=small}
\caption{List of the selected papers tackling classification task.}
\label{tab:classification}
\setlength{\tabcolsep}{0.8mm}{}
\renewcommand\arraystretch{2.0}
\begin{small}
\scalebox{0.7}{
\begin{tabular}{cccccc}
\toprule

\textbf{Task} & \textbf{Method} & \textbf{Year} & \textbf{Components} & \textbf{Evaluation} & \textbf{Dataset} \\

\midrule

\multirow{5}{*}{\rotatebox{90}{\textbf{TMI}}}
& \makecell[c]{TrajectoryNet~\cite{jiang2017trajectorynet}}~\href{https://github.com/wuhaotju/TrajectoryNet}{Code} & 2017 & GRU & \makecell[c]{Accuracy, \\CE Loss, F1 Score} & Geolife \\
\cline{2-6}
& \makecell[c]{ST-GRU~\cite{liu2019spatio}} & 2019 & GRU & Accuracy & \makecell[c]{Geolife, \\SH Taxi, \\Synthetic} \\
\cline{2-6}
& \makecell[c]{TrajODE~\cite{liang2021modeling}} & 2021 & RNN, ODE & Accuracy & \makecell[c]{Geolife, \\Grab-Posisi} \\
\cline{2-6}
& \makecell[c]{TraClets~\cite{kontopoulos2022traclets}~\href{https://github.com/AntonisMakris/d-LOOK}{Code}} & 2022 & CNN, FC & Accuracy & \makecell[c]{GeoLife, \\Hurricane, \\Animals} \\
\cline{2-6}
& \makecell[c]{TrajFormer~\cite{liang2022trajformer}~\href{https://github.com/yoshall/TrajFormer}{Code}} & 2022 & Transformer & \makecell[c]{Accuracy, \\FLOPs} & \makecell[c]{Geolife, \\Grab-Posisi} \\

\midrule

\multirow{5}{*}{\rotatebox{90}{\textbf{TUL}}}
& \makecell[c]{TULER~\cite{gao2017identifying}~\href{https://github.com/gcooq/TUL}{Code}} & 2017 & RNNs & \makecell[c]{Acc@k,
Macro-F1} & \makecell[c]{Gowalla, \\Brightkite} \\
\cline{2-6}
& \makecell[c]{TULVAE~\cite{zhou2018trajectory}~\href{https://github.com/AI-World/IJCAI-TULVAE/tree/master}{Code}} & 2018 & LSTM, VAEs & \makecell[c]{Acc@k, Macro-P, \\Macro-R, Macro-F1} & \makecell[c]{Gowalla, \\Brightkite, \\Foursquare} \\
\cline{2-6}
& \makecell[c]{DeepTUL~\cite{miao2020trajectory}~\href{https://github.com/CodyMiao/DeepTUL}{Code}} & 2020 & \makecell[c]{RNN \\ Attention} & \makecell[c]{Acc@k, Macro-P, \\Macro-R, Macro-F1} & \makecell[c]{Foursquare, \\WLAN} \\
\cline{2-6}
& \makecell[c]{MainTUL~\cite{chen2022MainTUL}~\href{https://github.com/Onedean/MainTUL}{Code}} & 2022 & \makecell[c]{LSTM \\ Attention} & \makecell[c]{Acc@k, Macro-P, \\Macro-R, Macro-F1} & \makecell[c]{Foursquare, \\Weeplaces} \\
\cline{2-6}
& \makecell[c]{AttnTUL~\cite{gao2020adversarial}~\href{https://github.com/Onedean/AttnTUL}{Code}} & 2023 & \makecell[c]{FC, GNN, \\Attention} & \makecell[c]{Acc@k, Macro-P, \\Macro-R, Macro-F1} & \makecell[c]{Private Car, \\Gowalla, \\Geolife} \\

\bottomrule
\end{tabular}

}
\vspace{-6mm}
\end{small}
\end{table}

\textbf{Trajectory-User Linking.} TUL associates anonymous semantic trajectories with specific users, facilitating applications such as epidemic tracking and personalized services. \textit{Formally, given an anonymous trajectory, TUL task aims to identify the actual user in the database corresponding to that journey}~\cite{chen2022MainTUL}. Key challenges include handling data sparsity and interpreting the hierarchical semantic structures inherent in human mobility.

TULER~\cite{gao2017identifying} pioneered this domain by using RNNs and word embeddings to link POI sequences to users, though it struggles with hierarchical semantics. To address this, TULVAE~\cite{zhou2018trajectory} incorporates variational autoencoders to manage sparsity and learn hierarchical features. DeepTUL~\cite{miao2020trajectory} further mitigates sparsity using attention mechanisms to capture multi-periodic patterns. More recent approaches include MainTUL~\cite{chen2022MainTUL}, which utilizes mutual distillation learning for temporal dependencies, AdattTUL~\cite{gao2020adversarial} employing GANs, and SML-TUL~\cite{zhou2021self}, which leverages contrastive learning under spatio-temporal constraints.

\textbf{Other perspectives.} Research also extends to semi-supervised and unsupervised settings. SECA~\cite{dabiri2019semi} and proxy-label methods~\cite{james2020semi} integrate labeled and unlabeled data, while SSFL~\cite{zhu2021semi} adapts this for federated learning. For unsupervised TMI, DeepCAE~\cite{markos2020unsupervised} combines convolutional autoencoders with clustering. Approaches for limited or unlabeled data include wavelet transformations~\cite{zhu2021improving}, map-matching integration~\cite{jiang2023framework}, and graph-based modeling in S2TUL~\cite{deng2023s2tul}. Furthermore, DPLink~\cite{feng2019dplink} and EgoMUIL~\cite{huang2023egomuil} address cross-platform heterogeneity, while AttnTUL~\cite{chen2023trajectory} employs hierarchical attention to handle varying trajectory densities.

\subsection{Travel Time Estimation}\label{subsec:estimation}

\textit{Travel Time Estimation (TTE), or Estimated Time of Arrival (ETA), is vital for location-based services, facilitating efficient trip management and route optimization~\cite{shen2025towards,wang2018will}. While traditional methods relying on origin-destination points often overlook path selection and road conditions, modern approaches integrate complex spatio-temporal data to enhance accuracy. These are primarily categorized into trajectory-based and road-based methods, as summarized in Table~\ref{tab:estimation}.}

\begin{table}[t!]
\centering
\captionsetup{font=small}
\caption{List of the selected papers tackling estimation task.}
\label{tab:estimation}
\setlength{\tabcolsep}{1.2mm}{}
\renewcommand\arraystretch{1.5}
\begin{small}
\scalebox{0.7}{
\begin{tabular}{ccccc}
\toprule

\textbf{Task} & \textbf{Method} & \textbf{Year} & \makecell[c]{\textbf{Components}\\(\textbf{Focus})}  & \textbf{Dataset} \\

\midrule

\multirow{4}{*}{\rotatebox{0}{\textbf{Trajectory}}}
& \makecell[c]{DeepTTE~\cite{wang2018will}}~\makecell[c]{\href{https://github.com/UrbComp/DeepTTE}{Code}} & 2018 & LSTM  & Geolife \\
& \makecell[c]{DeepTravel~\cite{zhang2018deeptravel}} & 2018 & BiLSTM   & \makecell[c]{Porto, Shanghai Taxi} \\
& \makecell[c]{MURAT~\cite{li2018multi}}~\makecell[c]{\href{https://github.com/liyaguang/deep-eta-murat}{Code}} & 2018 & \makecell[c]{Graph \\Embedding}  & \makecell[c]{NYC-Trip, BJS-Pickup} \\
& \makecell[c]{TTPNet~\cite{liang2022trajformer}}~\makecell[c]{\href{https://github.com/YibinShen/TTPNet}{Code}} & 2022 & RNN, GNN & \makecell[c]{Beijing Taxi, Shanghai Taxi} \\

\midrule

\multirow{5}{*}{\rotatebox{0}{\textbf{Road}}}
& \makecell[c]{WDR~\cite{wang2018learning}} & 2018 & LSTM, FC  & \makecell[c]{DiDi Beijing} \\
& \makecell[c]{DeepIST~\cite{fu2019deepist}}~\makecell[c]{\href{https://github.com/csiesheep/deepist}{Code}} & 2019 & PathCNN & \makecell[c]{Porto, \\Chengdu} \\
& \makecell[c]{ConSTGAT~\cite{fang2020constgat}} & 2020 & GAT  & \makecell[c]{Taiyuan, Hefei, HuiZhou} \\

& \makecell[c]{HetETA~\cite{hong2020heteta}}~\makecell[c]{\href{https://github.com/didi/heteta}{Code}} & 2020 & GCN & \makecell[c]{DiDi Shengyang} \\
& \makecell[c]{CompactETA~\cite{fu2020compacteta}} & 2020 & \makecell[c]{LSTM, FC} & \makecell[c]{Beijing, Suzhou, Shengyang} \\

\midrule

\multirow{8}{*}{\rotatebox{0}{\textbf{Others}}}
& \makecell[c]{ER-TTE~\cite{fang2021ssml}} & 2018 &  En route & \makecell[c]{Taiyuan, Hefei, HuiZhou} \\
& \makecell[c]{CatETA~\cite{ye2022cateta}} & 2022 & \makecell[c]{Classification}  & \makecell[c]{DiDi [Shenzhen,  Chengdu]} \\
& \makecell[c]{PP-TPU~\cite{liu2021privacy}} & 2021 & \makecell[c]{Uncertainty\\Privacy}  & \makecell[c]{Creteil, San Francisco} \\
& \makecell[c]{ProbTTE~\cite{liu2023uncertainty}} & 2023 & \makecell[c]{Classification\\Uncertainty}  & \makecell[c]{DiDi [Beijing, Shanghai]} \\
& \makecell[c]{DeepTTDE~\cite{james2021citywide}} & 2023 & \makecell[c]{Travel time\\
distributions}  & \makecell[c]{DiDi [Chengdu, Shenzhen]} \\

\bottomrule
\end{tabular}
}
\vspace{-4mm}
\end{small}
\end{table}

\textbf{Trajectory-based Estimation.} Leveraging trajectory sequences defined in Sec.~\ref{sec:defi_nota}, these methods predict travel time by analyzing GPS points. Wang~\textit{et al.} utilized raw GPS data via an error feedback recurrent convolutional neural network (eRCNN)~\cite{wang2016traffic}, while DeepTTE incorporated geo-convolution to capture spatial correlations~\cite{wang2018will}. Subsequent studies integrate grid-mapped trajectories with auxiliary data (\eg, traffic, weather) using multi-task learning and graph neural networks to enhance contextual analysis~\cite{zhang2018deeptravel, li2018multi, shen2020ttpnet, huang2022context}. Despite their utility, these methods suffer from reliance on high-frequency GPS data and the unrealistic assumption of known future locations. Consequently, road-based TTE has emerged as a robust alternative, mitigating sensitivity to sampling rates and positioning accuracy~\cite{ye2022cateta, wang2018learning}.

\textbf{Road-Based Estimation.} By defining trips as road sequences, these approaches model inter-road correlations to support diverse routing, thereby reducing trajectory dependence. WDR~\cite{wang2018learning} pioneered this using a hybrid regression framework to integrate comprehensive travel features. Subsequent research has incorporated metric learning and personalized driving behaviors to refine model sensitivity~\cite{sun2021road,sun2020codriver}, while PathCNN introduced sub-path images for spatio-temporal analysis~\cite{fu2019deepist}. Recently, the field has shifted towards complex networked representations, utilizing heterogeneous information graphs and spatio-temporal attention mechanisms to capture dynamic road contexts~\cite{hong2020heteta,fang2020constgat}. This evolution highlights the efficacy of advanced graph-based techniques in addressing road-based TTE challenges~\cite{fu2020compacteta,chen2022interpreting,jin2023dual,yuan2022route}.

\textbf{Other perspectives:} Beyond standard classifications, novel frameworks address TTE through alternative paradigms. Ye~\textit{et al.} and Liu~\textit{et al.} reframe TTE as a multi-classification task, mitigating long-tail effects by categorizing time spans based on trip distributions~\cite{ye2022cateta,liu2023uncertainty}. Fang~\textit{et al.} proposed en route TTE (ER-TTE), leveraging observed behaviors to adaptively refine predictions~\cite{fang2021ssml,fan2022metaer}. Furthermore, recent scholarship aims for robust, versatile solutions by exploring specialized domains, including uncertainty quantification~\cite{liu2021privacy,zhu2022cross}, cross-area generalization~\cite{zhu2022cross}, and travel time distribution estimation~\cite{zhou2023travel, james2021citywide}.

\subsection{Anomaly Detection}\label{subsec:detection}

\textit{Trajectory anomaly detection aims to identify abnormal movement of objects}, facilitating applications such as ride-hailing fraud detection, traffic monitoring, and trajectory cleaning. Early approaches, like TRAOD~\cite{trajanom_traod}, relied on hand-crafted distance-and-density rules to identify outliers. Methods are categorized into offline detection, which requires complete trajectories, and online detection, which supports ``on-the-fly'' processing. Notably, online methods offer superior flexibility by accommodating both real-time streams and progressively generated full trajectories.

\textbf{Offline Detection:} ATD-RNN~\cite{trajanom_rnn} utilizes an RNN with a fully connected layer to predict Euclidean anomalies via supervised learning. Addressing data scarcity, IGMM-GAN~\cite{trajanom_gan} employs an unsupervised CNN-based bidirectional GAN, where learned embeddings form a multi-modal Gaussian distribution, \ie~forming multiple clusters. Anomaly scores are derived from the distance between test trajectories and cluster centers. TripSafe~\cite{trajanom_tripsafe} targets ride-hailing anomalies by analyzing features like stopping duration, using dual VAEs to learn representations in both Euclidean and road network spaces. Furthermore, ATROM~\cite{gao2023open} applies variational Bayesian methods guided by probability measure rules to recognize anomalies in open-world scenarios.

\textbf{Online Detection:} DB-TOD~\cite{trajanom_dbtod} targets road networks by using reinforcement learning to model segment transition probabilities, formulating detection as a sequential decision process. RL4OASD~\cite{trajanom_subtraj} improves upon this by refining feature generation and introducing local rewards to enforce label continuity. Conversely, GM-VSAE~\cite{trajanom_gmvsae} operates in Euclidean space, adapting an RNN-based VAE to learn latent probability distributions and detect anomalies via generation likelihoods, thereby enhancing online efficiency. Building on this, DeepTEA\cite{trajanom_deeptea} further incorporates the temporal dimension into the detection framework.

\subsection{Mobility Generation}\label{subsec:generation}

\begin{figure}[htbp!]
 \centering
 \includegraphics[width=1.0\linewidth]{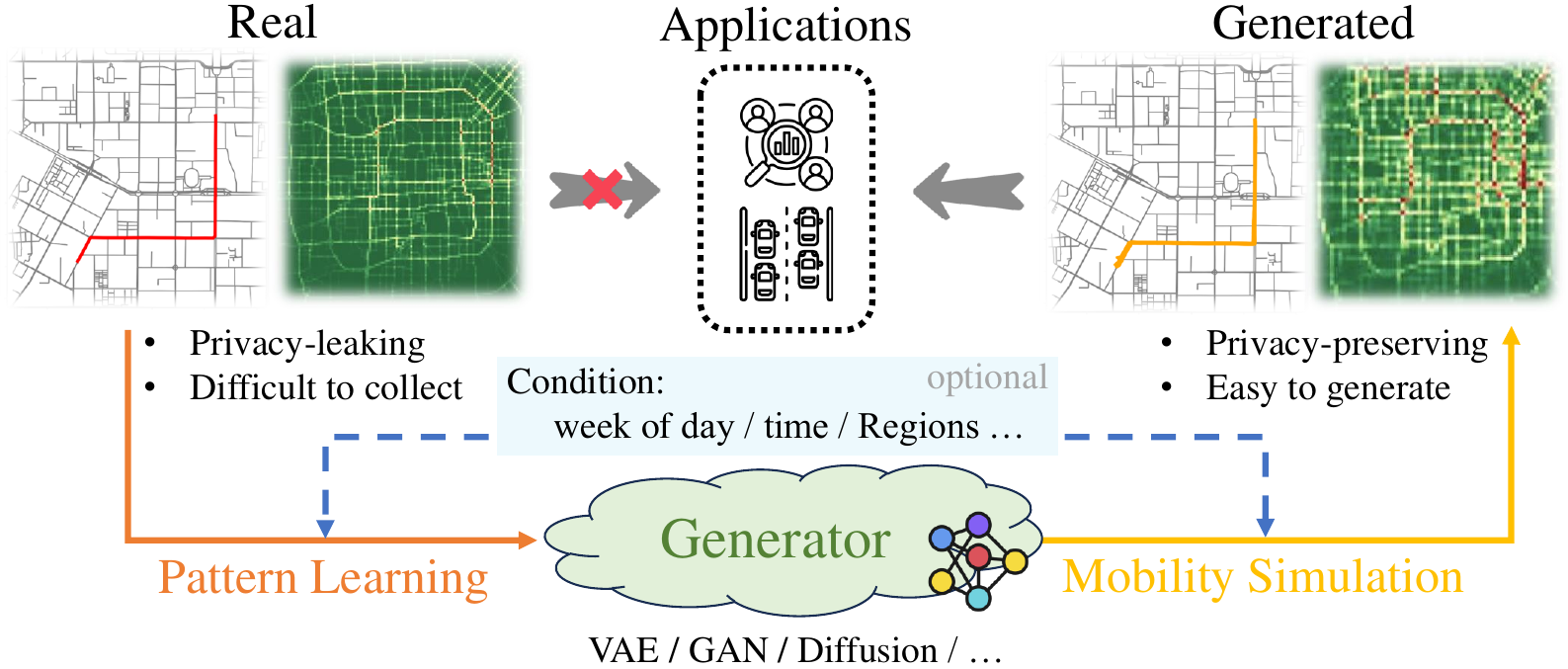}
 \caption{Macro and micro trajectory generation examples.}
 \label{fig:generation}
 \vspace{-2mm}
\end{figure}

Trajectory data applications are frequently hindered by data scarcity, privacy concerns, and authorization limits. To address these constraints, trajectory generation synthesizes realistic data that preserves privacy while supporting research utility~\cite{luca2021survey}. As illustrated in Figure~\ref{fig:generation}, \textit{trajectory generation employs deep learning to mimic complex movement patterns, encapsulating statistical, spatial, and temporal nuances to align generated outputs with real-world distributions.} Synthesis approaches are categorized by scale: macro-dynamics and micro-dynamics. Macro-dynamics model aggregate mobility trends, such as inter-regional population flows, emphasizing large-scale patterns over individual movements~\cite{hess2015data,luca2021survey}. Conversely, micro-dynamics focus on individual-level granularity—including specific routes, speeds, and stops—which is essential for high-resolution applications like location-based services and behavioral analysis.

\textbf{Macro-dynamics.}\label{subsec:gen_macro} Macro-level generation captures comprehensive mobility flows, traditionally utilizing statistical simulations and physics-based approaches like gravity and radiation models~\cite{barbosa2018human}. While foundational, these methods often oversimplify complex human dynamics. Recent deep learning advancements have transformed flow generation, employing architectures such as FC, CNN, RNN, and GANs to decode intricate spatiotemporal dependencies~\cite{chen2020citywide, wu2020spatiotemporal, zhang2019trafficgan, chen2019traffic, yin2018gans}. These data-driven methods significantly enhance flow fidelity and adaptability. Furthermore, graph-based models, such as the spatial interaction GCN proposed by Yao~\textit{et al.}, leverage local spatial networks to refine geographic unit representations~\cite{yao2018deep}.

\textbf{Micro-dynamics.}\label{subsec:gen_micro} Micro-level generation replicates individual mobility granularity, including location sequences, dwell times, and routes. Early approaches treated this as sequential next-location prediction, heavily relying on historical data~\cite{deeptransport, hong2022you,feng2018deepmove}, or utilized trajectory mixture models, which often entailed high computational overhead~\cite{nergiz2008towards}. The integration of Generative Adversarial Networks (GANs) introduced a paradigm shift; for instance, Ouyang~\textit{et al.} utilized grid-mapped data to synthesize realistic paths~\cite{ouyang2018non}, though trade-offs between grid resolution and accuracy persist~\cite{wang2021large, xu2021simulating}. Reinforcement learning further advanced the field by modeling trajectory generation as sequential decision-making processes~\cite{feng2020learning,wang2024cola,yuan2023learning}. Other methods transform trajectories into images, albeit with increased computational complexity~\cite{cao2021generating, wang2021large}. Most recently, denoising diffusion probabilistic models, such as DiffTraj~\cite{zhu2023difftraj} and Diff-RNTraj~\cite{wei2024diff}, have emerged, modeling generation via particle diffusion to achieve high-fidelity path creation.

\section{Recent Advances in Large Models for Trajectory Computing}

The rapid evolution of generative artificial intelligence, particularly large models demonstrating robust cross-modal reasoning and generalization, is revolutionizing trajectory computing. Traditional mining methods, typically reliant on task-specific deep learning models, suffer from poor generalizability and limited reasoning capabilities. Conversely, the emergence of \textbf{foundation models} (FM) and \textbf{large language models} (LLMs) offers a novel paradigm, facilitating more general, interpretable, and knowledge-integrated trajectory intelligence.

\subsection{Foundation Model in Trajectory Computing}

Since 2024, research has pivoted from city-specific architectures to unified foundation models capable of multimodal and cross-domain generalization~\cite{wu2025towards}. \textit{By employing Transformer and its variants as backbone networks to train on massive trajectory datasets from scratch, these models address traditional limitations regarding task specificity, regional dependence, data heterogeneity, and privacy preservation.} We categorize existing works by three key challenges.

\textbf{Geographic Scalability.} As conventional models use non-transferable spatial representations (grids/road IDs) tied to single cities, four transfer learning strategies have emerged. \textit{$\romannumeral1$) Region-agnostic Minimality:} UniTraj~\cite{zhu2024unitraj} utilizes the WorldTrace dataset (70 countries) via pure spatio-temporal points, excluding road networks and POIs. \textit{$\romannumeral2$) Unified Semantic Space:} MoveGPT~\cite{han2025trajmoe} embeds geography, POIs, and popularity, while UniMove~\cite{han2025unimove} adopts a feature-based trajectory-location dual-tower architecture. \textit{$\romannumeral3$) Unified Spatial Rendering:} VLMLocPredictor leverages VLM visual reasoning by rendering trajectories as images, enabling cross-city transfer. \textit{$\romannumeral4$) Transferable Location Encoding:} TrajFM~\cite{lin2024trajfm} incorporates POI modalities and learnable spatio-temporal rotary embeddings for vehicle trajectories.

\textbf{Handling Heterogeneity.} This involves managing heterogeneity in tasks (prediction, classification) and data (mixed patterns, noise). \textit{$\romannumeral1$) Task Heterogeneity:} TrajFM unifies tasks via ``trajectory mask-and-restore,'' while BIGCity~\cite{yu2025bigcity} uses spatio-temporal prompts to guide a frozen model across analyses without fine-tuning. \textit{$\romannumeral2$) Data Pattern Heterogeneity:} MoveGPT~\cite{han2025trajmoe} and UniMove~\cite{han2025unimove} utilize Mixture-of-Experts (MoE) architectures, implementing spatial-aware and mobility-aware routing~\cite{yuan2025breaking}, respectively. \textit{$\romannumeral3$) Data Quality Heterogeneity:} UniTraj~\cite{zhu2024unitraj} employs adaptive resampling and self-supervised masking to mitigate sampling rate inconsistencies.

\textbf{Data Issues.} To address data fragmentation caused by privacy regulations, MoveGCL~\cite{yuan2025breaking} introduces Generative Continual Learning. By using a frozen teacher model for experience replay and knowledge distillation, it prevents catastrophic forgetting, facilitating decentralized, privacy-preserving model evolution.

\subsection{Large Language Model for Trajectory Computing}

Since 2023, the integration of Large Language Models (LLMs) has fundamentally transformed trajectory computing. As trajectory data are inherently sequential and context-rich, they align well with LLM capabilities in sequence modeling and reasoning. Research has expanded beyond traditional deep learning to exploit LLMs for complex management and mining tasks, focusing on: \textit{$\romannumeral1$)} Domain adaptation via parameter tuning; \textit{$\romannumeral2$)} Multimodal semantic fusion; \textit{$\romannumeral3$)} Agentic frameworks for planning; and \textit{$\romannumeral5$)} Semantic benchmarking.

\textbf{LLM Fine-tuning and Alignment for Trajectory Data}.
Despite the potential of zero-shot approaches, \textit{Fine-tuning and Alignment} remain essential for domain specialization.
\begin{itemize}[leftmargin=*]
  \item \emph{Task-Specific Fine-tuning:} PLMTrajRec~\cite{wei2024plmtrajrec} addresses trajectory recovery by encoding sampling intervals into natural language prompts, enhancing generalization across rates. Similarly, Traj-LLM~\cite{lan2024traj} validates the adaptability of LLMs (\eg, GPT-2) for trajectory prediction in \textbf{few-shot} contexts via LoRA.
  \item \emph{Human Behavior Alignment:} Liu et al.~\cite{liu2025aligning} propose a framework to align LLM travel choices with human behavior without computationally intensive fine-tuning. It utilizes socio-demographic behavioral embeddings to construct persona loading functions, dynamically selecting appropriate persona-prompts for contextualized simulation.
\end{itemize}

\textbf{Multimodal and Semantic Fusion of Trajectory Data}.
Feeding non-textual trajectory data into LLMs requires advanced representation, encoding, and modality translation.
\begin{itemize}[leftmargin=*]
  \item \emph{Multimodal Trajectory Representation:} Traj-MLLM~\cite{liu2025traj} introduces a training-free framework that converts trajectories into interleaved image-text sequences via map-anchored tokenization. OmniTraj~\cite{zhu2025learning} unifies trajectory, topology, road segment, and regional semantics into a shared space to facilitate flexible retrieval and multimodal learning.
  \item \emph{Efficient Temporal Tokenization:} Addressing long-sequence inefficiencies, RHYTHM~\cite{he2025rhythm} implements Hierarchical Temporal Tokenization. By segmenting trajectories into daily units and applying hierarchical attention, it significantly reduces sequence length for efficient prediction using a frozen LLM backbone.
  \item \emph{Semantic Feature Extraction and Alignment:} IMPEL~\cite{nie2025joint} utilizes LLMs as geospatial knowledge encoders to generate transferable node representations for Spatio-Temporal Graph Neural Networks. TrajCogn~\cite{zhou2024trajcogn} aligns continuous spatio-temporal features with ``anchor word'' embeddings (\eg, ``turn,'' ``accelerate''), enabling the LLM to comprehend motion patterns and travel purposes.
\end{itemize}

\textbf{LLM-based Agentic Frameworks}.
Research increasingly positions LLMs as high-level \textit{``controllers''} orchestrating planning and reasoning, while specialized tools handle computation.
\begin{itemize}[leftmargin=*]
  \item \emph{Unified Modeling and Automation:} TrajAgent~\cite{dutrajagent} establishes a \textit{``large-and-small model collaboration''} paradigm. The LLM acts as a manager within a Unified Execution Environment, planning and executing tasks via specialized smaller models and optimizing performance through cooperative learning.
  \item \emph{Zero-Shot Prediction and Reasoning:} AgentMove~\cite{feng2025agentmove} tackles generalization in \textit{zero-shot next-location prediction} by coordinating three modules: Spatial-Temporal Memory (individual patterns), World Knowledge Generator (urban structure), and Collective Knowledge Extractor (group patterns), synthesizing these for final reasoning.
  \item \emph{Large-Scale Traffic Simulation:} To address scalability, MobiVerse~\cite{liu2025mobiverse} combines a lightweight generator for activity chains with an LLM-based modifier that reacts to dynamic environments (\eg, road closures), simulating over 50,000 agents in real-time. CAMS~\cite{du2025cams} aligns synthetic trajectories from a CityGPT-based agent with real-world data via \textit{Direct Preference Optimization}.
\end{itemize}

\textbf{Semantic Understanding and Benchmarking}.
Evaluating LLMs' semantic comprehension of trajectories involves specialized benchmarking. MobQA~\cite{asano2025mobqa} introduces a dataset testing three cognitive levels: 1) \textit{Factual Retrieval}, 2) \textit{Multiple-Choice Reasoning}, and 3) \textit{Free-Form Explanation}. Results indicate that while LLMs excel at factual retrieval, their capacity for deep semantic reasoning diminishes significantly with increasing trajectory sequence length.

\section{Application and Resources}\label{sec:app}

\vspace{-1mm}
\subsection{Application}\label{subsec:application}

\begin{figure}[htbp!]
    \centering
    \vspace{-2mm}
    \includegraphics[width=0.9\linewidth]{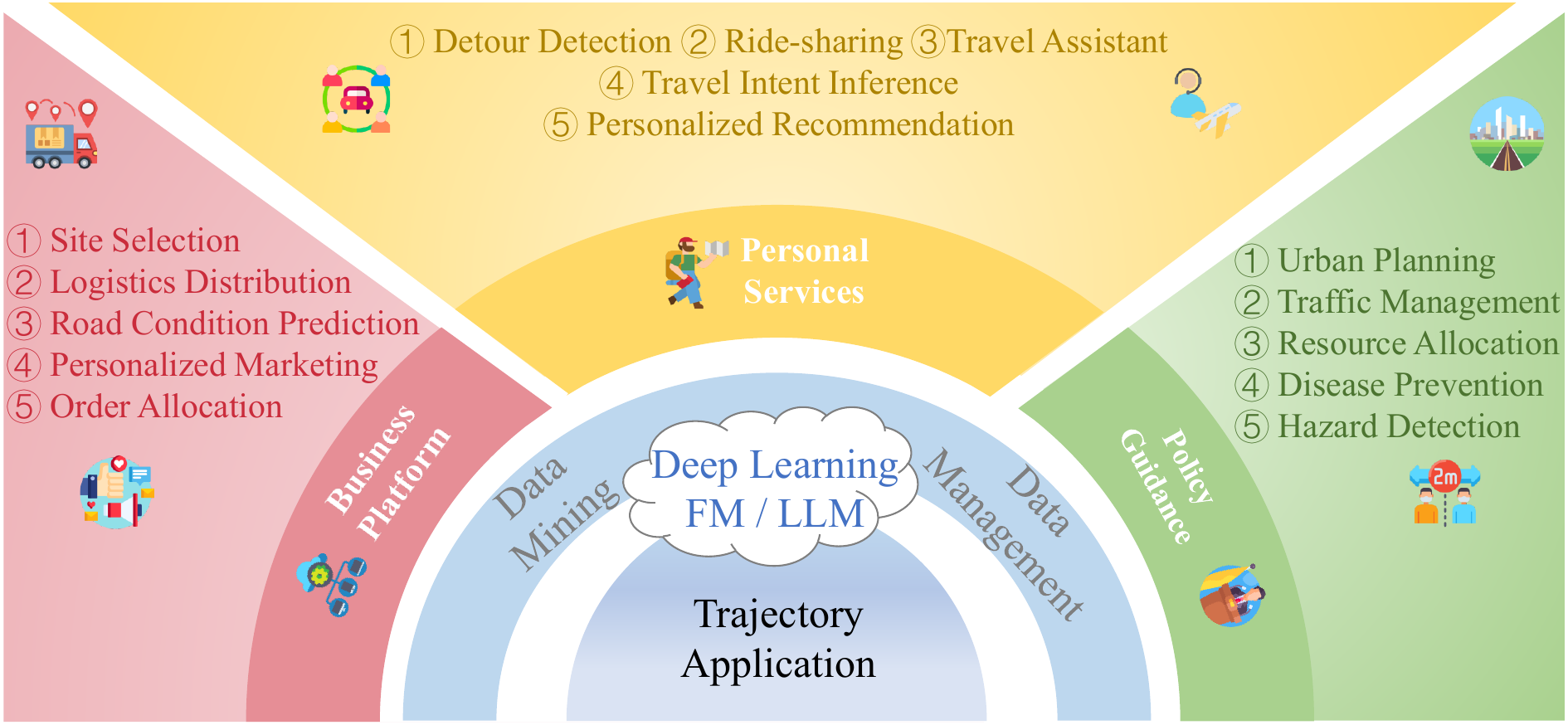}
    \caption{Trajectory application in various fields.}
    \label{fig:application}
    \vspace{-4mm}
\end{figure}

Trajectory data management and mining have revolutionary applications in various fields. As shown in Figure~\ref{fig:application}, we summarize these applications from different groups.

\textbf{Personal Services.}
Trajectory computing plays a vital role in various aspects of personal outdoor services. Firstly, in the aspect of route detection~\cite{zhang2023online}, the analysis of user's driving trajectories enables timely identification and notification of alternative routes or avoidance of traffic congestion, thereby enhancing travel efficiency. Secondly, ride-sharing~\cite{ma2013t} benefits from the application of trajectory data, as platforms can intelligently match passengers traveling in the same direction, leading to more efficient shared rides, reduced travel costs, and alleviated traffic burdens. Furthermore, personalized recommendation services~\cite{limcn4rec} utilize trajectory data analysis to understand users' preferred locations and behavioral patterns, delivering more tailored recommendations for nearby attractions, restaurants and business areas. Moreover, by combining semantics and multi-modal information, it can further analyze user travel intentions and serve as an intelligent agent~\cite{xie2024travelplanner} to assist users in decision-making.

\textbf{Business Platforms.}
Trajectory computing significantly influences business operations across various domains, especially for mobility service providers, such as Uber\footnote{\url{https://www.uber.com}}, DiDi\footnote{\url{https://didiglobal.com}}, Google Map\footnote{\url{https://www.google.com/maps}}, Baidu Map\footnote{\url{https://map.baidu.com}}, Tomtom\footnote{\url{https://www.tomtom.com}}, Cainiao\footnote{\url{https://www.cainiao.com}} and so on. In terms of business site selection~\cite{zou2017innovative}, the analysis of potential customers' movement trajectories empowers businesses to make informed decisions about optimal operational locations, thereby increasing the likelihood of business success. Additionally, logistics and delivery services benefit from trajectory data, enabling real-time monitoring and rational route planning to enhance delivery efficiency and reduce operational costs~\cite{wu2023lade}. Personalized marketing strategies leverage trajectory data analysis to understand user behavior, implementing more individualized marketing approaches to increase user engagement~\cite{wang2019efficiently}. Moreover, road condition prediction and travel order allocation~\cite{wen2023survey}, facilitated by real-time analysis, provide businesses with more accurate and efficient services, ultimately elevating overall operational standards.

\begin{table*}[t!]
	\centering
    \setlength{\tabcolsep}{1.0mm}{}
    \renewcommand\arraystretch{1.15}
	\caption{Publicly available trajectory datasets.}
	\vspace{-2mm}
	\label{tab:traj_dataset}
	\scalebox{0.765}{\begin{tabular}{llllllllllll}
                
                \toprule
			\textbf{Categorization} &\textbf{Type} & \textbf{Dataset Name} & \textbf{Main Area} & \textbf{Duration} & \textbf{Statistics} & \textbf{\#Point/Records} & \textbf{\#Attributes}\\

                \midrule
   
			\multicolumn{1}{c}{\multirow{17}{*}{\makecell[c]{Continuous\\GPS traces}}} 
    
                & Human & GeoLife~\cite{zheng2009mining}:~\href{https://www.microsoft.com/en-us/download/details.aspx?id=52367}{link} 
                    & Asia & 4.5 Years & 182 users, 17,621 trajectories, 91\% 1$\sim$5 s/p sample rate & 24.87 million+ & 7 \\
                & Human & TMD:~\href{https://cs.unibo.it/projects/us-tm2017/index.html}{link}     
                    & Italiana & 31 Hours & 13 users, 226 trajectories, 0.05 s/p sample rate & -- & 9 \\
                & Human & SHL:~\href{http://www.shl-dataset.org/}{link} 
                    & U.K. & 7 Months & 3 users, 12 trajectories, 1 s/p sample rate & -- & 28 \\
                & Human & OpenStreetMap:~\href{https://www.openstreetmap.org/traces}{link}
                    & Global & From 2005 & 8.7 million+ trajectories, continuously updating & -- & 7 \\
                & Human & MDC:~\href{https://www.idiap.ch/en/scientific-research/data/mdc}{link}
                    & Switzerland & 3 Years &  185 trajectories, nearly 200 individuals & 4,527,539 & -- \\
                
                & Taxi & T-Drive~\cite{yuan2011driving}:~\href{https://www.microsoft.com/en-us/research/publication/t-drive-trajectory-data-sample/}{link} 
                    & Beijing, China & 1 Weeks & 10357 cars, 177 s/p (Avg.) sample rate & 15 million+ & 4 \\
                & Taxi & Porto:~\href{https://www.kaggle.com/c/pkdd-15-predict-taxi-service-trajectory-i/data}{link} 
                    & Porto, Portugal & 9 Months & 442 cars, 1,710,990 trajectories, 15 s/p sample rate & 1,710,990 & 9 \\
                & Taxi & Taxi-Shanghai:~\href{https://cse.hkust.edu.hk/scrg}{link}
                    & Shanghai, China & 1 Year & 4,316 cars, 7.8 million trajectories, 5 s/p sample rate & -- & 5 \\
                & Taxi & DiDi-Chengdu 
                    & Chengdu, China & 1 Month & 3,493,918 trajectories, 3 s/p Avg. sample rate & 1.4 billion+ & 5 \\
                & Taxi & DiDi-Xi’an 
                    & Xi’an, China & 1 Month & 2,180,348 trajectories, 3 s/p Avg. sample rate & 1 billion+ & 5 \\
                & Car & WorldTrace:~\href{https://huggingface.co/datasets/OpenTrace/WorldTrace}{link} 
                & Global & 2 Years & 70 Countries, 2.45 million trajectories, 1 s/p sample rate & 880 million & 9+ \\
                & Truck & Greek:~\href{http://isl.cs.unipi.gr/db/projects/rtreeportal/}{link}
                    & Athens, Greece & -- & 50 trucks, 1,100 trajectories & 112,203 & 9 \\
                & Hurricane & HURDAT:~\href{https://www.aoml.noaa.gov/hrd/hurdat/Data_Storm.html}{link}     
                    & Atlantic & 151 Years & 1,415 trajectories, 6 h/p sample rate & -- & 5 \\
                & Delivery & \makecell[l]{Grab-Posisi-L~\cite{xu2020grab}} 
                    & Southeast Asia & 1 Months & 84K trajectories, 1 s/p sample rate & 80 million+ & 9 \\

                & Vehicle & NGSIM:~\href{https://catalog.data.gov/dataset/next-generation-simulation-ngsim-vehicle-trajectories-and-supporting-data}{link}
                    & USA & 45 Minutes & 0.1 s/p sample rate, collected through video cameras & -- & 20+ \\
                & Animal & Movebank:~\href{https://www.movebank.org/cms/movebank-main}{link}
                    & Global & Decades & 8,480 studies, 1,383 taxa, 4,139 data owners & 6.1 billion & -- \\
                & Vessel & Vessel Traffic:~\href{https://marinecadastre.gov/AIS/}{link}
                    & USA & 9 Years & 60s/p, AIS data & -- & 7+ \\

                \midrule

                \multicolumn{1}{c}{\multirow{14}{*}{\makecell[c]{Check-in\\sequences}}} 
                & Human & Gowalla:~\href{https://snap.stanford.edu/data/loc-gowalla.html}{link} 
                    & Global & 1.75 Years & 196,591 nodes, 950,327 edges & 6.44 million+ & 5 \\
                & Human & Brightkite:~\href{https://snap.stanford.edu/data/locbrightkite.html}{link}  
                    & Global & 30 Months & 58,228 nodes, 214,078 edges & 4,491,143 & 5 \\
                & Human & Foursquare-NY:~\href{https://sites.google.com/site/yangdingqi/home/foursquare-dataset}{link}    
                    & New York, USA & 10 Months & 38,336 venues, 824 users & 227,428 & 8 \\
                & Human & Foursquare-TKY:~\href{https://sites.google.com/site/yangdingqi/home/foursquare-dataset}{link}
                    & Tokyo, Japan & 10 Months & 61,858 venues, 1,939 users & 573,703 & 8 \\
                & Human & Foursquare-Global:~\href{https://sites.google.com/site/yangdingqi/home/foursquare-dataset}{link}
                    & Global & 18 Months & 3,680,126 venues, 266,909 users & 33,278,683 & 15  \\
                & Human & Weeplace:~\href{https://www.yongliu.org/datasets/}{link}
                    & Global & 7.7 Years & 971,309 venues, 15,799 users & 7,658,368 & 7 \\
                & Human & Yelp:~\href{https://www.yelp.com/dataset/}{link}
                    & Global & 15 Years & 131,930 venues, 1,987,897 users & 6,990,280 & 20+ \\
                & Human & Instagram~\cite{chang2018content-instagram} 
                    &  New York, USA & 5.5 Years & 13,187 venues, 78,233 users & 2,216,631 & -- \\
                & Human & GMove~\cite{zhang2016gmove} 
                    & 2 cities in USA & 20 Days & 72K trajectories & 1.3 million & -- \\
                
                & Taxi & TLC:~\href{https://www.nyc.gov/site/tlc/about/tlc-trip-record-data.page}{link}
                    & New York, USA & From 2009 & 115,990 vehicles & -- & 10+ \\
                
                & Bicycle & Mobike-Shanghai 
                    & Shanghai, China & 2 Weeks & 390K+ bikes & 60 million+ & 10 \\
                & Bicycle & Bike-Xiamen:~\href{https://data.xm.gov.cn/contest-series/digit-china-2021}{link}
                    & Xiamen, China & 5 Days & 50K+ bikes & 198,382 & 6 \\
                & Bicycle & Citi Bikes:~\href{https://citibikenyc.com/system-data}{link}
                    & New York, USA & From 2013 & 68K+ bikes, 2,104 active stations & 60K+/month & 13 \\
                    
                & Delivery & LaDe~\cite{wu2023lade}:~\href{https://wenhaomin.github.io/LaDe-website/}{link} 
                    & 5 cities in China & 6 Months & 21,000 users, 10,677,000 trajectories & -- & 17 \\

                \midrule
                \multicolumn{1}{c}{\multirow{2}{*}{\makecell[c]{Synthetic\\traces}}} 
                & Taxi & SynMob~\cite{zhu2023synmob}:~\href{https://yasoz.github.io/SynMob/}{link} 
                    & \makecell[l]{Chengdu, China\\Xi'an, China} & 1 Month & \makecell[l]{2,000,000 (unrestricted) trajectories,\\3 s/p Avg. sample rate}  & 1 billion+ & 4 \\
                & Vehicle & BerlinMod:~\href{https://secondo-database.github.io/BerlinMOD/BerlinMOD.html}{link}    
                    &  Berlin, German & 28 Days & 2,000 vehicles, 292,940 trajectories & 56,129,943 & -- \\

                \midrule

                \multicolumn{1}{c}{\multirow{10}{*}{\makecell[c]{Other formats\\of trajectories}}} 
                & Crowd Flow & COVID19USFlows:~\href{https://github.com/GeoDS/COVID19USFlows}{link} 
                    & USA & From 2019 & 220k venues, millions of anonymous users & -- & -- \\
                & Crowd Flow & MIT-Humob2023:~\href{https://connection.mit.edu/humob-challenge-2023}{link} 
                    & Japan & 90 Days &  \makecell[l]{100,000 individuals, 85 types of venues,\\30-minute intervals, 500-meter grid cells}  & -- & -- \\
                & Crowd Flow & BousaiCrowd:~\href{https://github.com/deepkashiwa20/DeepCrowd}{link} 
                    & Japan & 4 Months & 1 million users, 20 record/day sample rate & 150 million & 4 \\
                & Traffic Flow & TaxiBJ~\cite{zhang2017deep}:~\href{https://github.com/amirkhango/DeepST}{link} 
                    & Beijing, China & 17 Months & \makecell[l]{$32\times32$ grids,  60 s/p sample rate,\\30-minute intervals, 34,000+ taxis}& -- & -- \\ 
                 & Traffic Flow & BikeNYC~\cite{zhang2017deep}:~\href{https://github.com/amirkhango/DeepST}{link} 
                    & New York, USA & 6 Months & \makecell[l]{$16\times8$ grids,  60 s/p sample rate,\\1-hour intervals, 6,800+ bikes}& -- & -- \\ 
                & Traffic Flow & TaxiBJ21~\cite{jiang2022taxibj21}:~\href{https://github.com/jwwthu/DL4Traffic/tree/main/TaxiBJ21}{link} 
                    & Beijing, China & 3 Months & \makecell[l]{$32\times32$ grids, 600-meter cell length,\\30-minute intervals, 17,749 taxis}& -- & -- \\ 
                \bottomrule
	\end{tabular}}
	\vspace{-4mm}
\end{table*}

\textbf{Policy Guidance.}
Trajectory computing offers valuable insights for policymakers and urban planners. In terms of traffic management, trajectory data analysis allows for the intelligent adjustment of traffic signals~\cite{ma2020multi} and the rational planning of traffic flow~\cite{dai2015personalized}, thereby improving urban traffic efficiency. Urban planners can utilize trajectory data to provide a more accurate foundation for city planning~\cite{zheng2015trajectory} by understanding the activity trajectories of city residents, facilitating the scientific planning of urban infrastructure and land use. Resource allocation and disease control~\cite{levitt2020predicting} benefit from trajectory data mining, enabling governments to allocate urban resources more accurately and respond promptly to different regional needs. Real-time monitoring of human movement dynamics assists in the early detection of potential disease spread risks. Finally, in the realm of danger (crime) detection~\cite{wu2018location}, the application of trajectory data aids in identifying abnormal behavior, enhancing urban security and enabling law enforcement agencies to intervene and prevent potential criminal events more effectively.

\subsection{Resources}\label{subsec:resources}


Trajectory computing is crucial for understanding human mobility, and significant datasets and tools have been accumulated. We conduct a comprehensive analysis to address the current lack of a detailed survey of available open-source data and tools crucial for fostering transparent research.


\textbf{Datasets.}
Table~\ref{tab:traj_dataset} lists all known publicly available trajectory datasets, categorized into three groups based on the form of data collection: continuous GPS traces, check-in sequence, and synthetic traces.
The table encapsulates pertinent details for each dataset, including its type, main area, duration and statistical information.


\textbf{Tools.}
For effective analysis and simulation, researchers have various tools at their disposal. 
SUMO\footnote{\url{https://eclipse.dev/sumo}}, an open-source traffic simulator, provides a comprehensive environment for traffic modeling. 
SafeGraph\footnote{\url{https://docs.safegraph.com/docs/welcome}} offers an academic platform with access to large, anonymous datasets for privacy-preserving analysis.
Cblab\footnote{\url{https://github.com/caradryanl/CityBrainLab}}, a toolkit for scalable traffic simulation, consists of CBEngine, CBData, and CBScenario, enabling efficient simulations and training of traffic policies for large-scale urban scenarios. PyTrack\footnote{\url{https://github.com/titoghose/PyTrack}} is a comprehensive tool that allows for the modeling of street networks, conducting topological and spatial analyses, and performing map-matching on GPS trajectories.
PyMove\footnote{\url{https://pymove.readthedocs.io/en/latest}} can be used for the processing and visualization of trajectories and other spatio-temporal data.
TransBigData\footnote{\url{https://transbigdata.readthedocs.io/}} is a Python package for analyzing transportation big data and offers a systematic method for processing trajectories.
Traja\footnote{\url{https://github.com/traja-team/traja}} is a toolkit for numerically characterizing and analyzing the trajectories of moving animals.
MovingPandas\footnote{\url{https://github.com/movingpandas/movingpandas}} provides generalized trajectory data structures and functions for movement data exploration and analysis.
Scikit-mobility\footnote{\url{https://github.com/scikit-mobility/scikit-mobility}} is a library designed for human mobility analysis, synthetic trajectory generation, and privacy risks assessment.
Tracktable\footnote{\url{https://github.com/sandialabs/tracktable}} is a set of Python and C++ libraries for the processing and analysis of trajectory. Yupi\footnote{\url{https://github.com/yupidevs/yupi}} is a set of tools designed for collecting, generating and processing trajectory data.

For detailed information and library access, please visit our \href{https://github.com/yoshall/Awesome-Trajectory-Computing}{official GitHub repository}, a central hub for leading advancements in trajectory computing, featuring research papers, benchmark datasets, and source codes.


\section{Challenges and Directions}\label{sec:future}



\subsection{Current Challenges}

\begin{figure}[t!]
    \vspace{-1mm}
    \centering
    \includegraphics[width=0.9\linewidth]{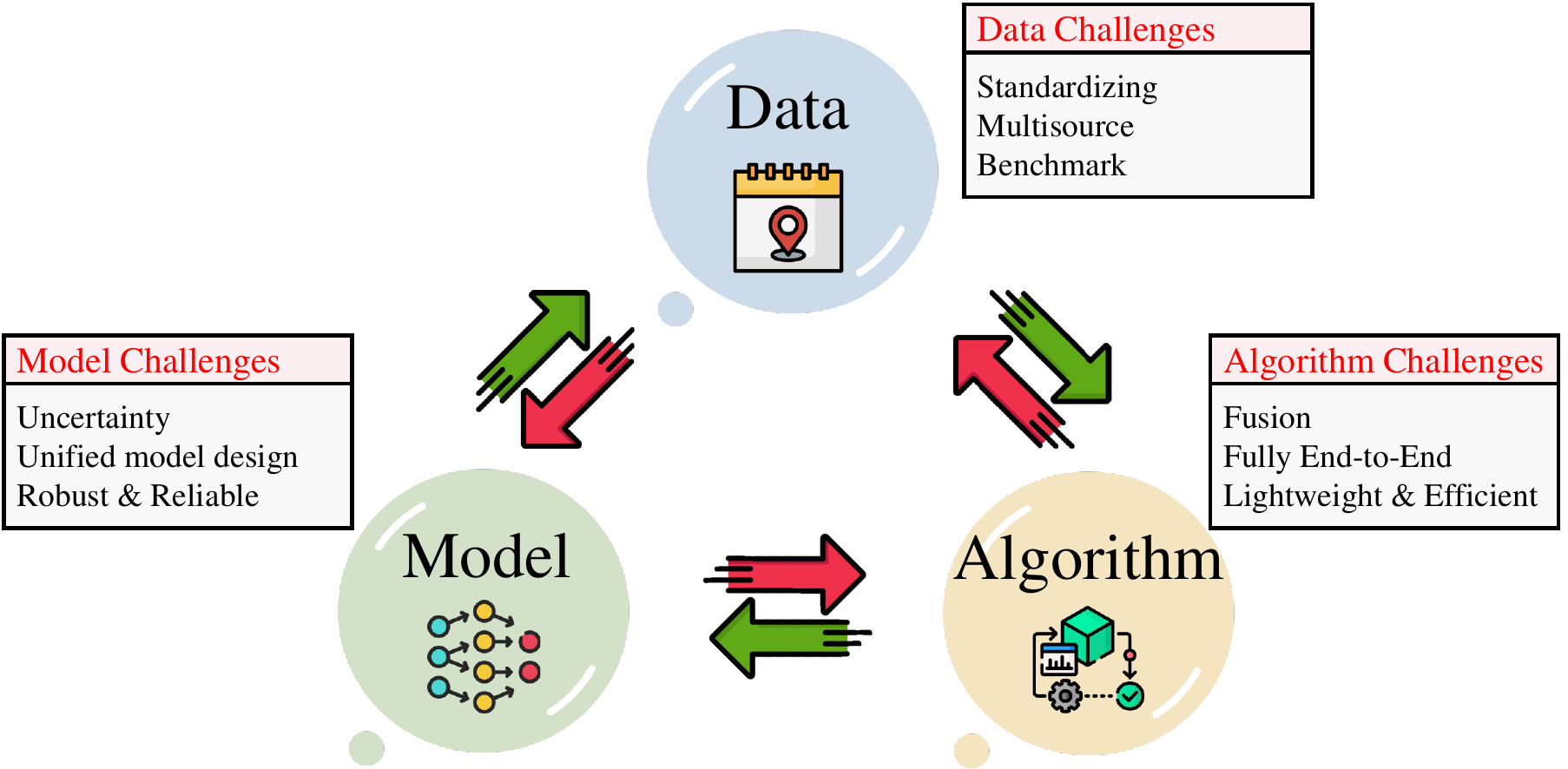}
    \caption{Current challenges facing the core triad.}
    \label{fig:challenges}
    \vspace{-6mm}
\end{figure}

Examining the core triad of data, models, and algorithms, we delineate the current status and challenges in Figure~\ref{fig:challenges}.

\textbf{Data.} \textit{$\romannumeral1$) Standardizing Trajectory Data Management:} Inadequate standardization impedes unified processing and application of trajectory data, necessitating open and standardized management approaches for seamless integration. \textit{$\romannumeral2$) Acquiring Multisource Semantic Trajectory Data:} Despite richer data from sources like social media, effective integration remains challenging. Advanced techniques are needed for acquiring and integrating diverse trajectory data to enhance deep learning models' multimodal understanding. \textit{$\romannumeral3$) Constructing Comprehensive Trajectory Datasets:} Large-scale, high-quality trajectory datasets are vital for deep learning model training. Balancing diversity and user privacy, along with ensuring spatio-temporal coverage, is crucial for improved model generalization.

\textbf{Model.} \textit{$\romannumeral1$) Modeling Uncertainty in Movement Behavior:} Handling uncertainty in trajectory data, with its sparse, noisy, and long-tailed distribution, requires robust models adaptable to real-world mobility complexities. \textit{$\romannumeral2$) Unified model design:} Specific model architecture hinders the exploration of unified patterns in trajectory data. It is particularly challenging to design unified models for different tasks. \textit{$\romannumeral3$) Robust, Reliable, and Stable Trajectory Modeling:} Existing models lack robustness in extreme outliers, especially in practical applications. Ensuring model reliability is imperative.

\textbf{Algorithm.} \textit{$\romannumeral1$) Fusion Algorithms for Multi-source Trajectory Data:} Existing algorithms for multi-source trajectory data can be more efficient. Robust algorithms are essential for global interpretative capabilities in fusing different data types. \textit{$\romannumeral2$) Fully End-to-End Algorithm Design:} Complete end-to-end algorithms simplify structures and enhance efficiency, addressing the multi-stage nature of current trajectory models. \textit{$\romannumeral3$) Lightweight and Efficient Algorithm Design:} Improving the efficiency of trajectory computing algorithms on resource-constrained edge devices is critical for practical applications.

\subsection{Future Directions}

Building upon the preceding analysis, we outline promising avenues for future research in deep learning for trajectory computing:

\textbf{Resolving Distribution Shifts.} Trajectory data exhibits significant spatiotemporal heterogeneity, causing distribution shifts between training and inference phases~\cite{ji2023self} that limit model generalization across diverse locations and time periods. Despite its critical impact, this challenge remains under-addressed in current architectural designs. Future research should investigate continual and incremental learning strategies to mitigate these shifts and enhance model robustness across varied datasets.

\textbf{Multi-Modality Fusion.} Human mobility is intrinsically linked to diverse data modalities—including visual, sensor, and textual data~\cite{zou2024deep,chen2024terra}—and distinct trajectory types such as taxi routes and public transit flows. As deep learning evolves towards unified multi-modal architectures~\cite{yan2023urban}, trajectory computing stands to benefit significantly. Future work must move beyond rudimentary concatenation to develop unified frameworks that effectively integrate heterogeneous data, thereby capturing comprehensive movement patterns and improving predictive accuracy.

\textbf{Foundation Models \& Large Language Models.} Current trajectory computing models often lack generality and external knowledge, relying heavily on task-specific scenarios. Foundation models and Large Language Models (LLMs)~\cite{jin2023spatio}, characterized by scalable parameters and compression capabilities, offer a pathway to unify trajectory tasks. While requiring careful cost-benefit optimization, integrating LLM knowledge is a rapidly emerging frontier. Key directions include distilling LLM knowledge to augment existing models and utilizing LLMs as autonomous decision-making agents~\cite{jin2024position}.

\textbf{Interpretability.} While deep learning in trajectory computing has prioritized performance through complex architectures, the interpretability of these black-box models remains largely unexplored. Identifying the causal factors driving predictive improvements is critical. Recent studies~\cite{luo2024towards} have begun incorporating causality and physical laws into network design to transcend mere statistical correlations. Consequently, developing interpretable, physics-informed, and causality-aware deep learning models represents a vital direction for achieving stable and robust predictions.

\textbf{Privacy and Security.} Addressing the privacy and security concerns inherent in trajectory data is imperative~\cite{jin2022survey,zhu2022cross}. Future research must focus on robust techniques for anonymization and sensitive information protection. Promising methodologies include the integration of federated learning for decentralized privacy preservation and the utilization of advanced generative models to synthesize high-fidelity, privacy-compliant trajectory data.

\vspace{-2mm}
\section{Conclusion}\label{sec:conclusion}
In this survey, we systematically explore the promising intersection between trajectory computing and deep learning (as well as recent large models). Our unified framework unveils a structured understanding of deep learning for trajectory computing, dissecting them into deep learning for trajectory data management and mining. This study offers a concise and organized perspective for researchers and practitioners. Examining existing methods, we provide fresh insights into the core contributions of deep learning to reshape trajectory computing and the field of mobility science, and summarize recent advancements in foundational and large language models in this direction. Furthermore, we summarize key application scenarios and resources, concluding with a discussion on open challenges and future research directions.






\bibliographystyle{IEEEtran}
\bibliography{IEEEabrv,Bibliography}




\end{document}